\PassOptionsToPackage{table}{xcolor}
\documentclass[]{antgroup}
\PassOptionsToPackage{numbers, compress}{natbib}
\usepackage{antgroup}

\usepackage{todonotes}
\usepackage{amsmath}
\usepackage{amssymb}
\usepackage{amsthm}
\usepackage{cleveref}
\usepackage{multicol}
\usepackage{array}
\usepackage{algorithm}
\usepackage{algorithmic}
\usepackage{tabularx}
\usepackage{longtable}
\usepackage{makecell}
\usepackage{mathtools}
\usepackage{amsfonts}
\usepackage{pifont}
\usepackage{tablefootnote}
\usepackage[normalem]{ulem}
\useunder{\uline}{\ul}{}
\usepackage{seqsplit}
\usepackage{xspace}
\usepackage{nicefrac}

\theoremstyle{plain}

\theoremstyle{definition}

\theoremstyle{remark}

\newcommand{\method}{\textsc{CausalMix}\xspace}

\newlength\savewidth
\newcommand\shline{\noalign{\global\savewidth\arrayrulewidth\global\arrayrulewidth .8pt}\hline\noalign{\global\arrayrulewidth\savewidth}}

\definecolor{firstcolor}{HTML}{C3423F}
\definecolor{secondcolor}{HTML}{2A4B8C}

\newcommand{\justificationTODO}[1][]{\textcolor{red}{\bf [TODO]}}

\title{\method: Data Mixture as Causal Inference for Language Model Training}

\author{
{\selectfont
Zinan Tang$^{1,2,\dagger,\ddagger}$~~
Yukun Zhang$^{2,\dagger}$~~
Shaomian Zheng$^{2}$~~
Zhuoshi Pan$^{1}$~~
Qizhi Pei$^{3}$\\
Dingnan Jin$^{2}$~~
Qing Cui$^{2}$~~
\textit{Jun Zhou}$^{2,\ast}$~~
\textit{Yujun Wang}$^{4,\ast}$}~~
\textit{Biqing Huang}$^{1,\ast}$
}

\affiliation[1]{Tsinghua University}
\affiliation[2]{Ant Group}
\affiliation[3]{Renmin University of China}
\affiliation[4]{Xiaomi Corporation}
\appto{\affiliationlist}{\par\href{mailto:tzn26@mails.tsinghua.edu.cn}{\texttt{tzn26@mails.tsinghua.edu.cn}}; \href{mailto:hbq@tsinghua.edu.cn}{\texttt{hbq@tsinghua.edu.cn}}}

\begin{document}

\maketitle

\renewcommand{\thefootnote}{\fnsymbol{footnote}}
\footnotetext[1]{Corresponding Authors.}
\footnotetext[2]{Equal Contribution.}
\footnotetext[3]{Work during research internship at Ant Group.}

\begin{abstract}

In Large Language Model (LLM) training, data mixing plays a pivotal role in determining model performance.
Recent methods optimize mixture weights via proxy models, but they rely on the assumption of static data distributions.
As a result, when the underlying data pool shifts, these methods require costly retraining from scratch.
This limitation restricts their ability to scale seamlessly from small settings to larger data pools and model sizes.
In this paper, we propose \method to address this limitation by casting data mixture optimization as a causal inference problem.
We formulate the statistical features of the data pool as covariates and the domain mixture as the treatment.
After fitting a causal model on 512 runs of \texttt{Qwen2.5-0.5B} to estimate the Conditional Average Treatment Effect (CATE), we extrapolate the optimal mixture for an 800K data pool and apply it to train a 7B model.
Furthermore, we successfully generalize the framework to long chain-of-thought data on \texttt{Qwen3-4B-Base}.
By leveraging causal modeling to isolate confounding biases, \method infers state-dependent data mixtures offline, with the selected mixture fixed during each training run.
Extensive experiments show that the mixture guided by \method achieves competitive performance across multiple downstream tasks compared with RegMix and other baselines. The CATE Interpreter visualizes the learned mixing strategy.
Overall, \method offers a causal and interpretable framework for optimizing LLM data mixtures.
\end{abstract}

\section{Introduction}
\label{sec:intro}

The remarkable capabilities of Large Language Models (LLMs) are driven by the quality and composition of their training data~\citep{zhang2025survey,kandpal2025position,tang-etal-2025-middo,gao-etal-2025-strategic}. During Supervised Fine-Tuning (SFT), where models are aligned with human intent and specialized for complex tasks, the data mixture, namely the relative proportion of different domains such as instruction following, mathematical reasoning, and coding, has a substantial impact on downstream performance~\citep{li2025data}. However, determining the optimal mixture remains a notoriously challenging problem. One reason is that training LLMs is expensive, making exhaustive grid search over the continuous simplex of mixture weights intractable for large-scale models.

Existing automated data mixing strategies typically approach this problem through the lens of representation learning or proxy modeling. Methods such as RegMix~\citep{liu2025regmix} optimize data weights by minimizing validation loss on a reference dataset, treating historical training runs as independent samples to fit a global mapping from mixture weights to loss. However, lower training loss does not necessarily imply better downstream performance~\citep{xu2026unveiling}. Moreover, global mappings fail to account for the profound impact of the \emph{data state}, namely the inherent complexity, quality, and difficulty of the specific data pool being used. In other words, a single mixture need not be optimal across different data states~\citep{wang2025tikmixdatainfluencedynamic,tao2026modalmix}.

To bridge this gap, we propose \method, a framework that formulates data mixture optimization not as a black-box hyperparameter search, but as a \emph{causal marginal return estimation problem}. Instead of seeking a universal mapping from mixture proportions $T$ to absolute performance $Y$, we treat historical proxy training runs as treatments. By conditioning on the data state $X$, characterized by metrics such as normalized loss~\citep{shum2025predictive}, entropy~\citep{li2026unified}, and writing style~\citep{10.5555/3692070.3694241}, we ask a localized causal question: \emph{How does a relative change in domain proportions causally affect downstream performance under the current data state?}

Drawing upon Double Machine Learning (DML)~\citep{chernozhukov2018double} and causal forests~\citep{wager2018estimation,oprescu2019orthogonal}, \method orthogonalizes the treatment and outcome variables with respect to the data state. This ensures that the estimated marginal returns are isolated from the confounding effects of the data pool's inherent quality. After identifying the causal direction, we update the mixture weights conservatively within a trust region.

The causal perspective of \method not only provides a principled optimization objective but also unlocks interpretability and transferability. By analyzing the Conditional Treatment Effects (CATE), we empirically unearth the ``skill conflicts'' between factual knowledge and complex logical reasoning~\citep{wu2025knowledge,balappanawar2025if}, and demonstrate how data quality thresholds dictate the effectiveness of math and coding data. Furthermore, because \method learns the underlying causal dynamics rather than memorizing a specific dataset, it successfully extrapolates to entirely unseen data pools and larger model architectures without requiring new proxy experiments. Taken together, these results position \method as a principled and practical framework for scalable, interpretable, and transferable data mixture optimization in LLM training.

\section{Methodology}
\label{sec:method}

\begin{figure}[!htb]
\centering
\includegraphics[width=\linewidth]{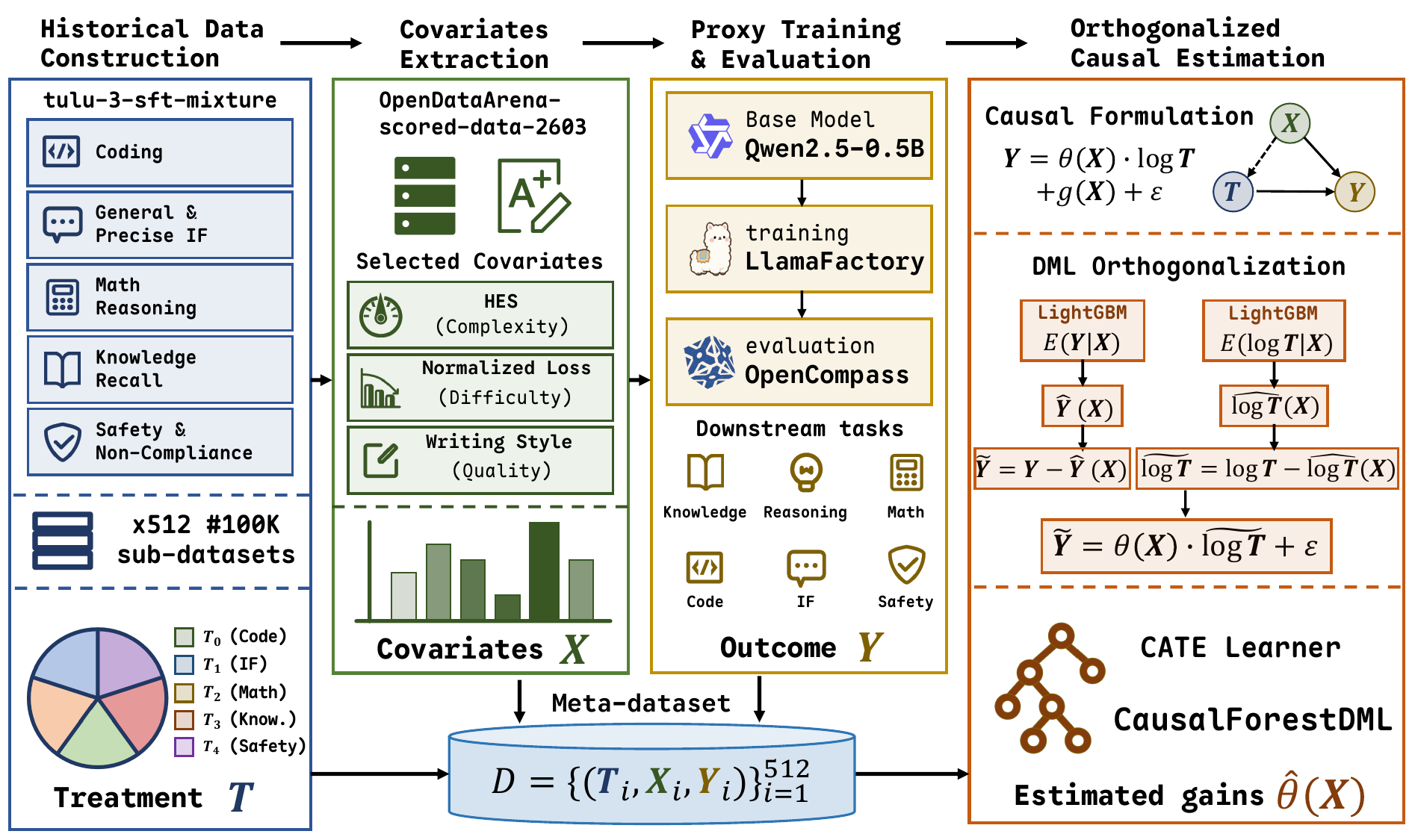}
\caption{Overview of the \method pipeline. Historical proxy runs provide data-state covariates, mixture assignments, and downstream outcomes, which are used to estimate state-conditioned marginal data returns through orthogonal causal learning.}
\label{fig:pipeline}
\end{figure}

We formulate data mixture optimization as a state-conditioned causal marginal return estimation problem. An overview of the \method pipeline is shown in Figure~\ref{fig:pipeline}.

\subsection{Target estimand and identification}
\label{sec:estimand_identification}

Given $K$ data domains and a fixed training budget, a mixture is represented as
\[
T=(T_1,\ldots,T_K), \qquad T_k\geq 0, \qquad \sum_{k=1}^{K}T_k=1.
\]
Each training run with a prescribed mixture can be viewed as a data-mixture treatment, and the resulting downstream performance is the corresponding outcome. To simultaneously capture the diminishing marginal returns dictated by empirical scaling laws~\citep{kaplan2020scaling} and accommodate the standard geometric transformations for compositional data on a probability simplex~\citep{aitchison1982statistical}, we define the continuous treatment using a log-mixture representation:
\[
Z=\log(T+\varepsilon),
\]
where the logarithm is applied element-wise and $\varepsilon>0$ is a small smoothing constant.

For the $i$-th historical proxy run, we first sample a candidate pool $P_i$ and compute covariates $X_i=\phi(P_i)$. We then assign mixture $T_i$, sample the final training set $D_i(T_i)$ from $P_i$ accordingly, and observe downstream performance $Y_i$ after training. The state $X_i$ describes the fixed pool's quality, difficulty, complexity, or stylistic statistics and remains unchanged when $T_i$ varies. Let $Y_i(t)$ be the potential outcome that would be obtained if run $i$ were trained with raw mixture $t$ from the same candidate pool $P_i$ under the same training budget, recipe, sampling rule, and evaluation protocol. We define the conditional response function for the log-mixture $z=\log(t+\varepsilon)$ as
\[
\mu(x,z)=\mathbb{E}[Y(t)\mid X=x].
\]
Learning the full response surface $\mu(x,z)$ is difficult because the mixture space is continuous and the number of proxy runs is limited. We therefore focus on the local marginal response within the treatment support covered by historical mixtures. For a given $x$, we use the partially linear approximation~\citep{chernozhukov2018double,robinson1988root,nie2021quasi}
\[
\mu(x,Z) \approx g(x)+\theta_0(x)^{\top}Z,
\]
where $g(x)$ captures the state-dependent baseline performance, and $\theta_0(x)\in\mathbb{R}^{K}$ is the \emph{state-conditioned marginal data return}. The quantity $\theta_0(x)$ can be understood as a generalized CATE for multidimensional continuous treatments. If $\theta_{0,k}(x)>0$, increasing the relative proportion of domain $k$ tends to improve downstream performance under state $x$; if $\theta_{0,k}(x)<0$, increasing that domain may induce negative transfer. Unlike feature importance in standard supervised learning, $\theta_0(x)$ describes the local causal response of potential outcomes to mixture treatments.

To identify this quantity from proxy runs, we assume consistency and ignorability by design:
\[
Y_i=Y_i(T_i), \qquad Y(t)\perp T\mid X.
\]
In our proxy design, mixtures are drawn independently following RegMix~\citep{liu2025regmix}, without using $P_i$, training feedback, or downstream outcomes. Since $Z$ is a deterministic transformation of $T$, this directly implies $Y(z)\perp Z\mid X$. We also assume local overlap and local smoothness within the historical treatment support. Under these assumptions,
\[
\mathbb{E}[Y\mid X=x,Z=z] = \mathbb{E}[Y(z)\mid X=x] = \mu(x,z),
\]
so the local marginal return $\theta_0(x)$ is identifiable from proxy mixture experiments. If mixtures are selected adaptively using training or evaluation results, this interpretation should be weakened to a causally motivated marginal-response estimate.

\subsection{Orthogonal estimation of marginal returns}
\label{sec:orthogonal_estimation}

Direct regression is not aligned with our objective, because it mixes state-dependent baseline effects with the causal effect of mixture changes. We address this issue using Double Machine Learning (DML)~\citep{chernozhukov2018double}, which residualizes both the outcome and the treatment with respect to the covariates to isolate the local causal response. We define the nuisance functions
\[
m_0(X)=\mathbb{E}[Y\mid X], \qquad e_0(X)=\mathbb{E}[Z\mid X],
\]
and construct residuals
\[
\widetilde{Y}=Y-m_0(X), \qquad \widetilde{Z}=Z-e_0(X).
\]
The marginal return is estimated from the residualized relation
\[
\widetilde{Y} \approx \theta_0(X)^{\top}\widetilde{Z}.
\]
This asks whether deviations in log-mixture proportions beyond their state-conditioned expectation explain performance deviations beyond the state-conditioned baseline.

In practice, the nuisance functions are estimated with cross-fitting~\citep{oprescu2019orthogonal,wager2018estimation}: historical proxy runs are split into folds, and each residual is generated by first-stage models trained without the corresponding sample. We then learn a heterogeneous effect model by minimizing the orthogonal loss
\[
\hat{\theta} = \arg\min_{\theta} \sum_i \left( \widetilde{Y}_i - \theta(X_i)^{\top}\widetilde{Z}_i \right)^2.
\]
This is an R-loss-style objective: it does not optimize absolute-score prediction, but instead estimates how residual treatment variation explains residual outcome variation.

\subsection{From marginal returns to mixture policies}
\label{sec:policy_extraction}

After estimating the target-state marginal return $\hat{\theta}(X_{\mathrm{tar}})$ with respect to log-mixture proportions, we convert it into a feasible raw mixture on the simplex. The guiding principle is simple: domains with larger positive log-mixture marginal returns should receive larger weights, while domains with low or negative marginal returns should not be encouraged to increase.

A deterministic analytical extraction maps positive log-mixture marginal returns to the simplex:
\[
T^{\mathrm{A}}_k = \frac{[\hat{\theta}_k(X_{\mathrm{tar}})]_+}{\sum_{j=1}^{K} [\hat{\theta}_j(X_{\mathrm{tar}})]_+}, \qquad [a]_+=\max(a,0).
\]
The detailed mathematical proof is provided in Appendix~\ref{app:analytical_proof}. And a search-based extraction instead evaluates a set of raw candidate mixtures $T^{(1)},\ldots,T^{(M)}$. Each candidate is transformed into its log-treatment representation $Z^{(m)}=\log(T^{(m)}+\varepsilon)$ before being passed to the fitted causal model. Let $\widehat{S}(Z^{(m)};X_{\mathrm{tar}})$ denote the predicted score or predicted gain of candidate $T^{(m)}$ at the target state. The final strategy is obtained by averaging the top candidates in the original mixture space:
\[
T^{\mathrm{S}} = \frac{1}{K_{\mathrm{top}}} \sum_{m\in\mathrm{Top}} T^{(m)}.
\]
This can be viewed as local bagging over high-scoring candidates: instead of relying on a single potentially overestimated mixture, it averages several strong raw-mixture candidates to reduce inference noise, smooth the resulting policy, and enhance generalization.

\section{Experiments}
\label{sec:experiments}

We first describe the experimental setup, then compare \method with strong baselines across different data scales and model sizes, further conduct extension experiments on LongCoT data, and finally present ablation studies. Appendix~\ref{app:exp_details} details the datasets, models, benchmarks, baselines, training and evaluation hyperparameters, and computing costs.

\subsection{Experimental setup}


\textbf{Data preparation}. We use the \texttt{tulu-3-sft-mixture}~\citep{lambert2025tulu} dataset and adopt the domain partitioning strategy introduced in Tulu 3. Specifically, we consider five domains: Coding, Instruction Following (IF, combining General and Precise IF), Math Reasoning, Knowledge Recall, and Safety \& Non-Compliance. For each of 512 proxy runs, we first sample a candidate pool $P_i$ and compute $X_i=\phi(P_i)$. Following RegMix~\citep{liu2025regmix}, we independently sample $T_i$ and use it to draw a 100K-example training set $D_i(T_i)$ from $P_i$. To efficiently extract data features, we leverage \texttt{OpenDataArena-scored-data-2603}~\citep{opendataarena_tool_2025,cai2025opendataarena,opendataarena_scored_data_2025}, which provides pre-computed scores on 30 metrics spanning multiple dimensions such as Diversity, Complexity and Quality. Three of these metrics, namely \texttt{Normalized\_Loss}~\citep{shum2025predictive}, \texttt{Writing\_Style}~\citep{10.5555/3692070.3694241}, and \texttt{HES}~\citep{li2026unified}, are aggregated over $P_i$ to form $X_i$. Section~\ref{sec:cov_impact} examines their impact.



\begin{table}[!htb]
\centering
\caption{Performance comparison of different data mixture methods across different model sizes and data scales. The highest scores are highlighted in bold, and the second-highest are underlined.}
\resizebox{\linewidth}{!}{
\begin{tabular}{lcccccc>{\columncolor{cyan!10}}c>{\columncolor{cyan!5}}c}

\shline
\textbf{Method} & \textbf{Knowledge} & \textbf{Reasoning} & \textbf{Math} & \textbf{Coding} & \textbf{IF} & \textbf{Safety} & $\mathbf{Avg}_{\mathrm{Dev}}$ & $\mathbf{Avg}_{\mathrm{Uns}}$ \\
\shline

Qwen2.5-0.5B-Instruct & 29.90 & 32.14 & 35.84 & 34.10 & 30.31 & 11.14 & 28.90 & 28.60 \\
Qwen2.5-7B-Instruct & 60.24 & 52.36 & 54.70 & 68.67 & 44.36 & 60.47 & 56.80 & 46.94 \\
Llama-3.1-Tulu-3-8B-SFT & 51.63 & 67.37 & 56.14 & 57.91 & 68.39 & 46.39 & 57.97 & 41.46 \\

\shline
\rowcolor{blue!15}\multicolumn{9}{c}{\textbf{\textit{Qwen2.5-0.5B, \# 100K, tulu-3-sft-mixture}}} \\
\shline
Equal & 27.65 & 29.80 & 23.32 & 33.00 & 33.83 & 14.57 & 27.03 & 24.43 \\
Grid & 28.46 & 31.98 & 30.41 & 33.50 & 20.15 & 23.38 & 27.98 & 23.45 \\
RegMix & 27.29 & 30.04 & 25.53 & 31.28 & 17.01 & 32.56 & 27.28 & \textbf{26.24} \\
DoReMi & 29.81 & 31.70 & 22.90 & 33.40 & 35.30 & 25.34 & \underline{29.74} & 24.12 \\
ODM & 27.50 & 30.53 & 25.41 & 33.10 & 26.06 & 23.62 & 27.70 & 23.59 \\
DMO & 28.86 & 30.91 & 22.51 & 30.68 & 35.86 & 26.19 & 29.17 & \underline{26.08} \\
\shline
\textbf{\method}-A & 28.37 & 31.31 & 24.67 & 29.57 & 38.63 & 26.93 & \textbf{29.91} & 23.42 \\
\textbf{\method}-S & 27.27 & 30.23 & 27.77 & 34.94 & 34.94 & 14.81 & 27.85 & 25.90 \\


\shline
\rowcolor{blue!15}\multicolumn{9}{c}{\textbf{\textit{Qwen2.5-0.5B, \# 400K, tulu-3-sft-mixture}}} \\
\shline
Equal & 28.39 & 32.98 & 23.96 & 32.10 & 37.89 & 27.78 & 30.51 & 24.71 \\
Grid & 29.34 & 31.84 & 26.68 & 35.02 & 16.64 & 17.26 & 26.13 & 27.08 \\
RegMix & 28.57 & 30.19 & 27.02 & 34.32 & 24.21 & 16.65 & 26.82 & \underline{26.07} \\
DoReMi & 28.58 & 31.12 & 22.14 & 33.61 & 39.74 & 26.56 & 30.29 & 25.69 \\
ODM & 28.65 & 30.74 & 24.35 & 34.62 & 30.50 & 12.61 & 26.91 & 25.51 \\
DMO & 27.74 & 31.75 & 24.25 & 31.80 & 38.63 & 41.62 & \underline{32.63} & 25.84 \\
\shline
\textbf{\method}-A & 28.73 & 30.63 & 24.67 & 30.48 & 42.51 & 43.45 & \textbf{33.41} & 24.26 \\
\textbf{\method}-S & 29.45 & 31.30 & 26.37 & 32.40 & 38.82 & 25.95 & 30.71 & \textbf{26.93} \\

\shline
\rowcolor{blue!15}\multicolumn{9}{c}{\textbf{\textit{Qwen2.5-0.5B, \# 800K, tulu-3-sft-mixture}}} \\
\shline
Equal & 28.07 & 29.28 & 21.59 & 36.06 & 46.95 & 25.09 & 31.78 & \underline{25.64} \\
Grid & 23.94 & 24.40 & 29.67 & 23.81 & 17.93 & 22.64 & 31.17 & 22.53 \\
RegMix & 24.90 & 29.94 & 30.25 & 25.46 & 25.32 & 22.52 & 26.40 & 22.82 \\
DoReMi & 27.70 & 29.66 & 20.58 & 34.53 & 42.88 & 32.93 & 31.38 & 24.88 \\
ODM & 28.59 & 30.50 & 25.18 & 36.06 & 33.64 & 10.28 & 27.37 & 25.36 \\
DMO & 28.07 & 31.88 & 22.42 & 26.90 & 41.59 & 41.37 & 32.04 & \textbf{26.49} \\
\shline
\textbf{\method}-A & 27.93 & 29.68 & 23.56 & 27.76 & 43.81 & 50.92 & \textbf{33.94} & 25.02 \\
\textbf{\method}-S & 28.31 & 30.96 & 27.64 & 30.97 & 42.51 & 36.47 & \underline{32.81} & 25.04 \\

\shline
\rowcolor{blue!15}\multicolumn{9}{c}{\textbf{\textit{Qwen2.5-7B, \# 800K, tulu-3-sft-mixture}}} \\
\shline
Equal & 60.85 & 64.55 & 59.03 & 53.61 & 68.58 & 53.49 & 60.02 & \textbf{49.55} \\
Grid & 59.08 & 68.77 & 65.99 & 61.55 & 44.92 & 56.43 & 59.46 & 46.37 \\
RegMix & 59.60 & 68.12 & 63.37 & 55.76 & 58.04 & 55.94 & 60.14 & 48.12 \\
DoReMi & 58.02 & 63.20 & 57.35 & 57.28 & 68.21 & 52.75 & 59.47 & 46.19 \\
ODM & 60.25 & 65.69 & 59.42 & 44.22 & 63.59 & 54.83 & 58.00 & 48.00 \\
DMO & 59.15 & 63.70 & 60.62 & 54.05 & 70.24 & 54.35 & 60.35 & 48.98 \\
\shline
\textbf{\method}-A & 57.14 & 64.03 & 58.51 & 65.52 & 68.21 & 57.65 & \underline{61.84} & \underline{49.09} \\
\textbf{\method}-S & 59.35 & 62.88 & 58.63 & 64.43 & 67.47 & 60.95 & \textbf{62.28} & 47.98 \\

\shline
\end{tabular}
}
\label{tab:main_result}
\end{table}

\textbf{Proxy model training and evaluation.} We select Qwen2.5-0.5B~\citep{qwen2.5,qwen2} as the proxy model and conduct training using \texttt{LlamaFactory}~\citep{zheng2024llamafactory}. For evaluation, we use \texttt{OpenCompass}~\citep{2023opencompass} to assess the models on a diverse suite of downstream tasks aligned with the training domains, following the Tulu 3 evaluation protocol~\citep{lambert2025tulu}. We group the downstream tasks into six capabilities: Knowledge, Reasoning, Math, Coding, IF and Safety. We further partition these benchmarks into Development set $\mathcal{S}_{\mathrm{Dev}}$ and Unseen set $\mathcal{S}_{\mathrm{Uns}}$. We adopt the domain-level micro-average score on $\mathcal{S}_{\mathrm{Dev}}$ as the final outcome $Y$.


\textbf{Causal model fitting and inference.} We use the \texttt{EconML}~\citep{econml} framework for causal model fitting and inference. Specifically, we adopt \texttt{LightGBM}~\citep{10.5555/3294996.3295074} as the first-stage predictor and \texttt{CausalForestDML}~\citep{wager2018estimation,chernozhukov2018double,oprescu2019orthogonal} as the core causal estimator; detailed rationales for these choices are provided in Section~\ref{sec:causal_select} and Section~\ref{sec:first_select}. After fitting the causal model on a meta-dataset of 512 historical $(X,T,Y)$ triplets, we set the covariate $X$ to the comprehensive feature profile of the full \texttt{tulu-3-sft-mixture} training dataset. We consider two variants: \method-A (Analytical), which directly computes the exact closed-form solution. And \method-S (Search), following the practice of RegMix~\citep{liu2025regmix}, we draw $100{,}000$ candidate mixtures from a Dirichlet distribution and perform inference on these candidates. We then average the top-100 performing mixtures to obtain the final strategy.


\textbf{Baselines.} We compare \method against several representative baselines. These include Grid, which denotes the best mixture proportion empirically identified from the 512 proxy-model runs, as well as existing automated mixing methods including RegMix~\citep{liu2025regmix}, DoReMi~\citep{xie2023doremi}, ODM~\citep{albalak2023efficientonlinedatamixing} and DMO~\citep{li2025data}. To ensure a fair comparison, rather than directly adopting the static mixture proportions reported in the original papers, we re-implement these automated methods and train them on our own historical runs following their official protocols. For DMO, we instead use the mixing ratios reported in its paper.

\subsection{Main result}


As illustrated in Table~\ref{tab:main_result}, \method achieves strong performance on $\mathbf{Avg}_{\mathrm{Dev}}$ and remains competitive on $\mathcal{S}_{\mathrm{Uns}}$. Notably, \method-S often performs better than \method-A on $\mathbf{Avg}_{\mathrm{Uns}}$, which may result from averaging the top-100 candidate mixtures: this procedure can smooth out idiosyncratic variance in individual solutions and thus yield a more robust strategy. To reduce the possibility that the observed gains are due to chance, we conduct repeated comparisons across multiple training data scales, ranging from 100K to 800K. Across these settings, our method remains competitive with strong baselines, including the recent SFT-oriented method DMO. Inspired by the \textit{rank invariance hypothesis} proposed by RegMix~\citep{liu2025regmix}, we further scale the model size to 7B under the 800K data setting and observe a similar performance trend. This cross-scale consistency supports our approach's effectiveness and robustness.

\subsection{Extension experiments}
\label{sec:extens_exp}


\begin{table}[!htb]
\centering
\caption{Performance comparison of different data mixture methods on LongCoT data. The highest scores are highlighted in bold, and the second-highest are underlined.}
\label{tab:extension}
\resizebox{0.8\linewidth}{!}{
\begin{tabular}{lcc>{\columncolor{cyan!5}}ccc>{\columncolor{cyan!5}}c>{\columncolor{cyan!10}}c}

\shline
\textbf{Method} & \textbf{GSM8K} & \textbf{MATH} & $\mathbf{Avg}_{\mathrm{Math}}$ & \textbf{HumanEval} & \textbf{MBPP} & $\mathbf{Avg}_{\mathrm{Code}}$ & $\mathbf{Avg}$ \\

\shline
\rowcolor{blue!15}\multicolumn{8}{c}{\textbf{\textit{Qwen3-4B, \# 20K, AM-Thinking-v1-Distilled-Code\&Math}}} \\
\shline

Equal & 90.45 & 56.78 & \underline{73.62} & 59.76 & 48.20 & 53.98 & \underline{63.80} \\
Grid & 87.34 & 61.20 & 74.27 & 62.80 & 47.60 & 55.20 & 64.74 \\
RegMix & 89.61 & 40.80 & 65.21 & 61.59 & 53.60 & 57.60 & 61.40 \\
DoReMi & 88.55 & 42.22 & 65.39 & 63.41 & 53.80 & \textbf{58.61} & 62.00 \\
ODM & 88.32 & 41.16 & 64.74 & 63.41 & 42.20 & 52.81 & 58.77\\
DMO & 89.61 & 54.38 & 72.00 & 54.88 & 55.00 & 54.94 & 63.47 \\

\shline
\textbf{\method} & 88.86 & 60.58 & \textbf{74.72} & 62.20 & 55.00 & \underline{58.60} & \textbf{66.66} \\
\shline
\end{tabular}
}
\end{table}

To rigorously evaluate the transferability of \method, we conduct an extended generalization experiment across disparate data pools and model architectures. Specifically, we repurpose the historical data from \texttt{tulu-3-sft-mixture}~\citep{lambert2025tulu}, retain the same covariates for $X$, and define the outcome $Y$ as the average downstream performance in the coding and math domains. Subsequently, we apply the trained causal predictor to the entirely unseen dataset \texttt{AM-Thinking-v1-Distilled-math\&code}~\citep{tian2025correctanswersequaldistillation} to infer the optimal mixture proportions. To validate the effectiveness of these extrapolated weights, we train and evaluate \texttt{Qwen3-4B}~\citep{qwen3technicalreport}, a model series distinct from the proxy model (the \texttt{Qwen2.5} series~\citep{qwen2}). Empirical evaluations demonstrate that \method consistently achieves the best performance. This robust transferability demonstrates that our \method successfully captures the intrinsic laws of data mixing, enabling seamless extrapolation across datasets and models without costly proxy-model retraining.

\subsection{Ablation study}

\begin{table}[!htb]
\centering
\caption{Ablation study of the key components in \method. Removing the DML orthogonalization step (\textit{w/o Orth.}) or discarding covariates (\textit{w/o} $X$) both lead to performance degradation. The highest average scores are highlighted in bold.}
\label{tab:ablation}
\resizebox{0.8\linewidth}{!}{
\begin{tabular}{lcccccc>{\columncolor{cyan!5}}c}

\shline
\textbf{Method} & \textbf{Knowledge} & \textbf{Reasoning} & \textbf{Math} & \textbf{Coding} & \textbf{IF} & \textbf{Safety} & $\mathbf{Avg}$ \\

\shline
\rowcolor{blue!15}\multicolumn{8}{c}{\textbf{\textit{Qwen2.5-0.5B, \# 800K, tulu-3-sft-mixture}}} \\
\shline

\textit{w/o $X$}  & 29.27 & 31.39 & 29.97 & 33.41 & 39.37 & 36.35 & 33.29 \\
\textit{w/o Orth.} & 27.41 & 31.29 & 24.74 & 31.90 & 41.04 & 37.82 & 32.66 \\

\shline
\textbf{\method}-A & 27.93 & 29.68 & 23.56 & 27.76 & 43.81 & 50.92 & \textbf{33.94} \\
\textbf{\method}-S & 28.31 & 30.96 & 27.64 & 30.97 & 42.51 & 36.47 & 32.81 \\
\shline

\rowcolor{blue!15}\multicolumn{8}{c}{\textbf{\textit{Qwen2.5-7B, \# 800K, tulu-3-sft-mixture}}} \\
\shline

\textit{w/o $X$} & 60.45 & 63.95 & 61.16 & 55.62 & 69.69 & 56.92 & 61.30 \\
\textit{w/o Orth.} & 59.50 & 64.66 & 60.45 & 45.82 & 68.76 & 58.75 & 59.65  \\

\shline
\textbf{\method}-A & 57.14 & 64.03 & 58.51 & 65.52 & 68.21 & 57.65 & 61.84 \\
\textbf{\method}-S & 59.35 & 62.88 & 58.63 & 64.43 & 67.47 & 60.95 & \textbf{62.28} \\
\shline

\end{tabular}
}
\end{table}

We compare \method against two degraded variants, both of which use \texttt{LightGBM} as the underlying regressor, to validate the necessity of its key components. (1) \textit{w/o $X$.} We entirely remove the state covariates, yielding a RegMix-like variant~\citep{liu2025regmix}. Unlike RegMix, however, its optimization target is not validation loss but the average performance on downstream tasks. In this setting, the model reduces to learning a global mapping from treatment to outcome, $\hat{Y} = g(T).$ By attempting to learn this static mapping without conditioning on the data state, this context-agnostic variant becomes highly vulnerable to distribution shifts, leading to the performance degradation. (2) \textit{w/o Orth.} We bypass the DML orthogonalization step and directly concatenate covariates $X$ and treatment $T$ to predict the absolute outcome $\hat{Y}$, i.e., $\hat{Y} = f(X, T).$
As shown in Table~\ref{tab:ablation}, this direct regression leads to clear performance degradation, even performing worse than directly fitting $T$, which further highlights the regularization bias inherent in standard supervised learning.

\section{Analysis}
\label{sec:analysis}


In this section, we analyze the choice of causal estimator and the impact of covariates. We further use the CATE model interpreter to provide interpretable insights into the dynamics of data mixing.

\subsection{Causal model selection}
\label{sec:causal_select}


To identify the most suitable causal estimator, we perform model selection using the R-Scorer (R-loss) metric. The R-Scorer provides a principled and approximately unbiased criterion based on\citet{robinson1988root}'s orthogonalization technique. It enables us to compare different causal estimators by evaluating how well their predicted causal effects $\hat{\theta}(X)$ explain variation in the residual outcomes $\tilde{Y}$ given the residual treatments $\tilde{T}$. We evaluate a range of causal estimators in the \texttt{EconML} framework that support multidimensional continuous treatments, and report the results in Table~\ref{tab:model_selection} (a). Among them, \texttt{CausalForestDML} achieves the best performance. We attribute this advantage to its non-parametric, tree-based recursive partitioning architecture. Unlike linear causal models that impose rigid parametric assumptions, \texttt{CausalForestDML} is suited to capturing the complex interactions between the multidimensional covariates and treatment space. It also naturally accommodates feature saturation and localized heterogeneous effects, making it suitable for the intricate dynamics of data mixing~\citep{wager2018estimation,chernozhukov2018double,oprescu2019orthogonal}.

\begin{table}[!htb]
    \centering
    \caption{Model selection results for the causal estimator and first-stage predictors. Left (a): performance of candidate causal estimators measured by RScore. Right (b): representative first-stage predictor combinations ranked by RScore. The selected models are highlighted by color.}
    \label{tab:model_selection}
    \begin{subtable}[t]{0.38\textwidth}
        \centering
        \resizebox{\linewidth}{!}{%
        \begin{tabular}{lc}
            \toprule
            \textbf{Model} & \textbf{RScore} ($\uparrow$) \\
            \midrule
            \texttt{LinearDML} & +0.1445 \\
            \texttt{SparseLinearDML} & -1.7065 \\
            \rowcolor{blue!15}\textbf{\texttt{CausalForestDML}} & \textbf{+0.1683} \\
            \texttt{CausalForestDML\_Deep} & -0.1238 \\
            \texttt{CausalForestDML\_Shallow} & +0.0207 \\
            \texttt{DML\_Poly2\_Lasso} & +0.1404 \\
            \texttt{DML\_Poly2\_Ridge} & +0.0340 \\
            \texttt{DML\_Poly3\_Lasso} & +0.1533 \\
            \texttt{DML\_Poly3\_Ridge} & +0.0653 \\
            \bottomrule
        \end{tabular}%
        }
    \end{subtable}
    \hfill
    \begin{subtable}[t]{0.61\textwidth}
        \centering
        \resizebox{\linewidth}{!}{%
        \begin{tabular}{ccccc}
            \toprule
            \textbf{Rank} & $\mathbf{Y}$ \textbf{Predictor} & $\mathbf{T}$ \textbf{Predictor} & \textbf{RScore} ($\uparrow$) & \textbf{Time} (s) \\
            \midrule
            \rowcolor{blue!15}1 & \textbf{\texttt{LightGBM}} & \textbf{\texttt{LightGBM}} & \textbf{0.1683} & 12.9 \\
            5 & \texttt{RandomForest} & \texttt{LightGBM} & -0.0556 & 10.3 \\
            6 & \texttt{RidgeCV} & \texttt{LightGBM} & -0.1408 & 6.8 \\
            7 & \texttt{ElasticNetCV} & \texttt{LightGBM} & -0.1681 & \textbf{5.5} \\
            8 & \texttt{GradientBoosting} & \texttt{LightGBM} & -0.1686 & 15.4 \\
            19 & \texttt{RandomForest} & \texttt{RandomForest} & -0.2840 & 18.7 \\
            23 & \texttt{GradientBoosting} & \texttt{RandomForest} & -0.3241 & 22.5 \\
            24 & \texttt{LassoCV} & \texttt{LightGBM} & -0.4098 & 11.1 \\
            25 & \texttt{LightGBM} & \texttt{RandomForest} & -0.4323 & 18.0 \\
            29 & \texttt{GradientBoosting} & \texttt{GradientBoosting} & -0.4816 & 10.0 \\
            \bottomrule
        \end{tabular}%
        }
    \end{subtable}
\end{table}

\subsection{First-stage predictor selection}
\label{sec:first_select}

The first-stage predictors estimate the conditional expectations $\hat{Y}(X)$ and $\hat{T}(X)$. We evaluate a diverse set of regression algorithms and their combinations for the outcome and treatment models. As summarized in Table~\ref{tab:model_selection} (b), with the full results provided in the Appendix, using \texttt{LightGBM} for both models achieves the highest RScore. This configuration substantially outperforms all other standalone regressors as well as linear models. Notably, \texttt{LightGBM} also emerges as the best treatment predictor across all top-ranked configurations. We attribute this strong performance to \texttt{LightGBM}'s efficient framework, which handles the multidimensional statistical features while capturing variable interactions~\citep{10.5555/3294996.3295074}. Although its computational cost is not the lowest among the candidates, this trade-off is acceptable given the substantial performance gains.

\subsection{Impact of covariates}
\label{sec:cov_impact}


To examine the impact of covariates, we compare metrics from \texttt{\seqsplit{OpenDataArena-scored-data-2603}}~\citep{opendataarena_scored_data_2025} using 64 randomly sampled instances from our 512 historical records as a validation set. Keeping all other hyperparameters fixed, we train causal models with different covariate combinations and evaluate predictions using Spearman rank correlation with ground-truth scores. We compare combinations across different sizes.

\begin{figure}[!htb]
    \centering
    \includegraphics[width=0.65\linewidth]{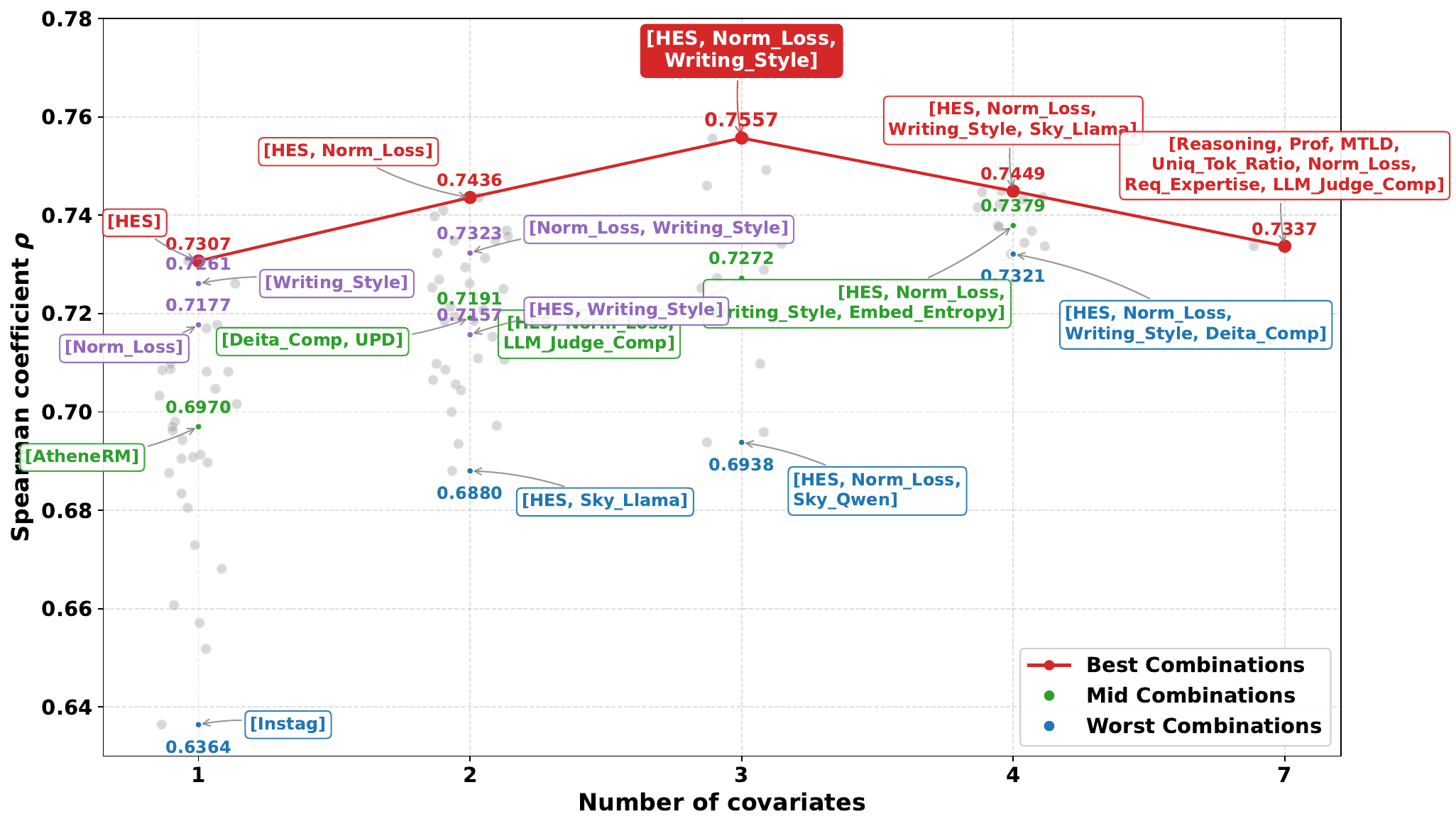}
    \caption{Spearman rank correlation under different covariate combinations.}
    \label{fig:x_spearman}
\end{figure}

Figure~\ref{fig:x_spearman} shows that performance varies across covariate combinations, with three covariates yielding the highest rank correlation. Specifically, \texttt{HES} sums the entropy of the top 0.5\% highest-entropy tokens in reasoning traces produced by \texttt{Qwen3-8B} to capture critical decision points and genuine reasoning complexity~\citep{li2026unified}. \texttt{Normalized\_Loss} computes the normalized cross-entropy using \texttt{Qwen3-8B}~\citep{shum2025predictive}, reflecting data predictability and training utility. Finally, \texttt{Writing\_Style} evaluates the clarity, coherence, and stylistic quality of the text using \texttt{QuRater-1.3B}~\citep{10.5555/3692070.3694241}.

These three metrics reflect data Complexity, Difficulty, and Quality, respectively, suggesting that complementary data properties help model mixture effects. Adding more covariates does not necessarily improve performance, possibly due to the curse of dimensionality with limited proxy records. More proxy runs may support richer covariate representations.

\subsection{CATE Interpreter}

\begin{figure}[!htb]
    \centering
    \includegraphics[width=0.8\linewidth]{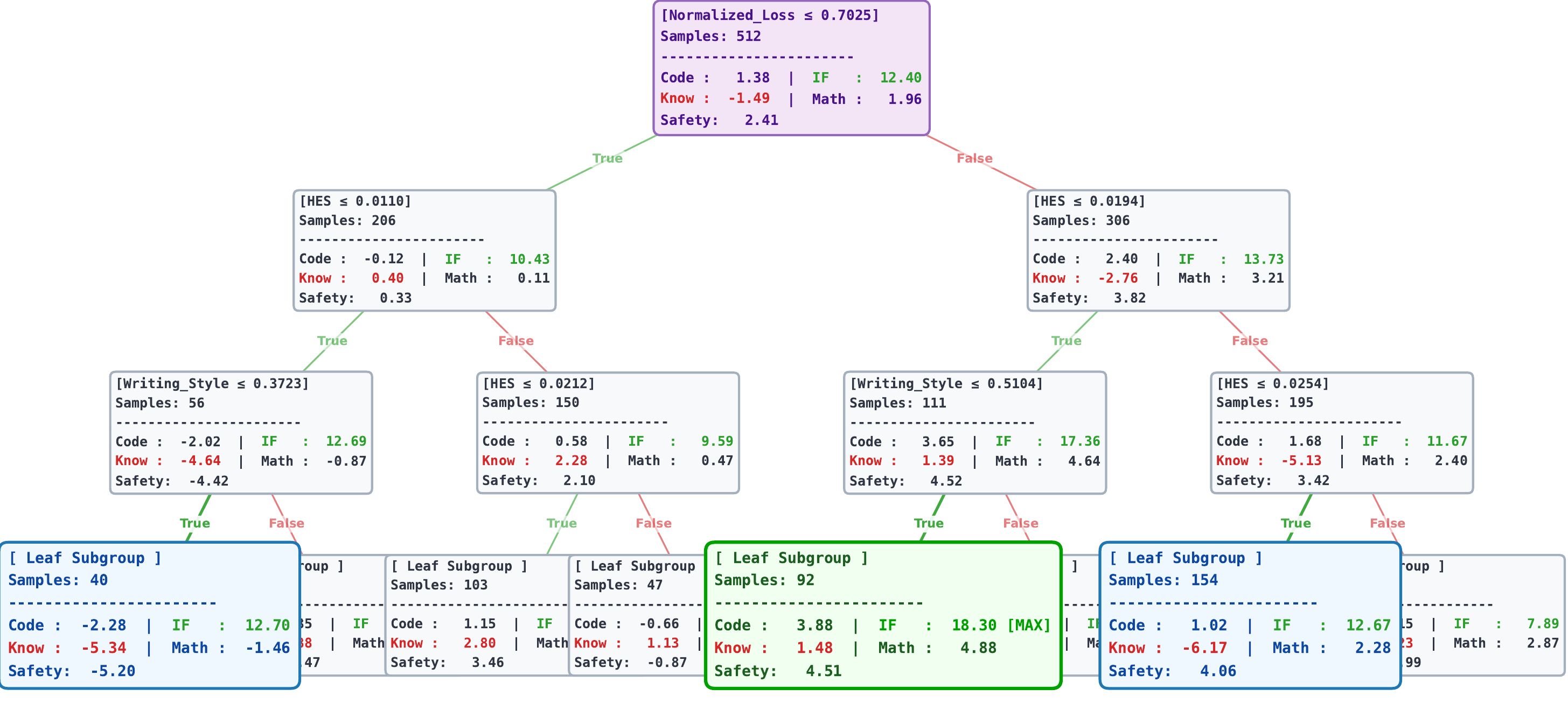}
    \caption{Simplified visualization of the CATE model tree interpreter.}
    \label{fig:cate_tree}
\end{figure}

We conduct a Tree Interpreter analysis of the trained causal model, as shown in Figure~\ref{fig:cate_tree}. The results show that IF data is the primary driver of downstream alignment, yielding stable positive returns across feature subspaces. In contrast, Knowledge data has negative effects on difficult target data characterized by high \texttt{Normalized\_Loss} and high \texttt{HES}, corroborating the existence of ``skill conflicts'' between logical reasoning and factual knowledge injection~\citep{wu2025knowledge,balappanawar2025if}. Moreover, the marginal returns of different domains depend strongly on the characteristics of the target data. In low-quality regions, characterized by low \texttt{Writing\_Style} and low \texttt{HES}, complex domains such as Math, Coding, and Safety introduce distributional noise and degrade performance. However, when \texttt{Writing\_Style} and \texttt{HES} are moderate, these domains produce strong synergistic gains, mitigating the performance penalty typically associated with Safety data.

\section{Related works}
\label{sec:related}

\paragraph{Data mixture optimization.}

Data mixture plays an important role in LLM training and strongly affects downstream task performance. Most existing offline methods~\citep{xie2023doremi,albalak2023efficientonlinedatamixing,liu2025regmix,fan2024doge,ye2025data,chen2025aioli} focus on the pre-training stage, deriving domain weights through proxy models or modeling training loss as a function of the data mixture. In contrast, data mixture optimization for SFT remains relatively underexplored. Existing SFT-oriented methods, such as DMO~\citep{li2025data} and IDEAL~\citep{ming2026ideal}, still fundamentally use validation loss as the optimization objective. SMART~\citep{renduchintala-etal-2024-smart} is a relatively rare exception that does not directly optimize validation loss; instead, it formulates data selection as two consecutive cardinality-constrained submodular maximization problems.

\paragraph{Causal inference in machine learning.}


Integrating causal inference with machine learning helps mitigate spurious correlations and distribution shifts in traditional data-driven models~\citep{peters2017elements,scholkopf2021toward}. This line of research is grounded in the potential outcomes framework~\citep{rubin2005causal,imbens2015causal} and causal graphical models~\citep{pearl2009causality,spirtes2000causation}. Recent work has mainly progressed along three directions: improving treatment effect estimation through deep representation learning for confounder control~\citep{shalit2017estimating,louizos2017causal,shi2019adapting} and DML frameworks~\citep{chernozhukov2018double}; uncovering latent data-generating structure through differentiable causal discovery~\citep{zheng2018dags} and mechanism disentanglement~\citep{bengiometa}; and improving generalization by incorporating causal invariance into objectives~\citep{rojas2018invariant,arjovsky2019invariant,liu2021towards}.

\section{Conclusion}
\label{sec:conclusion}

In this work, we introduced \method, a framework that shifts SFT data mixture optimization from static validation-loss minimization to state-conditioned causal marginal return estimation. By treating historical proxy training runs as causal treatments and combining orthogonalized estimation with a conservative trust-region policy, \method isolates the marginal utility of domain proportions from the confounding effects of the underlying data state. Extensive experiments show that our approach remains competitive across different model scales and data budgets, while also exhibiting strong transferability to unseen LongCoT data pools. Furthermore, the interpretable insights derived from our causal framework, including quantified skill conflicts between factual knowledge injection and complex logical reasoning, provide a principled foundation for future research on understanding and optimizing the dynamics of LLM training. Explicitly modeling model-dependent marginal returns and adapting mixtures during training are left to future work.

\section*{Acknowledgements}

This work was supported by Ant Group Research Intern Program and the National Science and Technology Major Project of China under grant 2025ZD1404600. We also thank Dr.~Gang Wang of Xiaomi Corporation for his support.

\bibliographystyle{antgroup}
\bibliography{references}

@article{nie2021quasi,
  title={Quasi-oracle estimation of heterogeneous treatment effects},
  author={Nie, Xinkun and Wager, Stefan},
  journal={Biometrika},
  volume={108},
  number={2},
  pages={299--319},
  year={2021},
  publisher={Oxford University Press}
}

@article{aitchison1982statistical,
  title={The statistical analysis of compositional data},
  author={Aitchison, John},
  journal={Journal of the Royal Statistical Society: Series B (Methodological)},
  volume={44},
  number={2},
  pages={139--160},
  year={1982},
  publisher={Wiley Online Library}
}

@misc{qwq32b,
    title = {{QwQ-32B}: Embracing the Power of Reinforcement Learning},
    url = {https://qwenlm.github.io/blog/qwq-32b/},
    author={{Qwen Team}},
    month = {March},
    year = {2025}
}

@misc{ji2025amthinkingv1advancingfrontierreasoning,
      title={AM-Thinking-v1: Advancing the Frontier of Reasoning at 32B Scale}, 
      author={Yunjie Ji and Xiaoyu Tian and Sitong Zhao and Haotian Wang and Shuaiting Chen and Yiping Peng and Han Zhao and Xiangang Li},
      year={2025},
      eprint={2505.08311},
      archivePrefix={arXiv},
      primaryClass={cs.CL},
      url={https://arxiv.org/abs/2505.08311}, 
}

@article{10.1145/3654979,
author = {Li, Peng and He, Yeye and Yashar, Dror and Cui, Weiwei and Ge, Song and Zhang, Haidong and Rifinski Fainman, Danielle and Zhang, Dongmei and Chaudhuri, Surajit},
title = {Table-GPT: Table Fine-tuned GPT for Diverse Table Tasks},
year = {2024},
issue_date = {June 2024},
publisher = {Association for Computing Machinery},
address = {New York, NY, USA},
volume = {2},
number = {3},
url = {https://doi.org/10.1145/3654979},
doi = {10.1145/3654979},
journal = {Proc. ACM Manag. Data},
month = may,
articleno = {176},
numpages = {28}
}

@inproceedings{
han2024wildguard,
title={WildGuard: Open One-stop Moderation Tools for Safety Risks, Jailbreaks, and Refusals of {LLM}s},
author={Seungju Han and Kavel Rao and Allyson Ettinger and Liwei Jiang and Bill Yuchen Lin and Nathan Lambert and Yejin Choi and Nouha Dziri},
booktitle={The Thirty-eight Conference on Neural Information Processing Systems Datasets and Benchmarks Track},
year={2024},
url={https://openreview.net/forum?id=Ich4tv4202}
}

@inproceedings{
wildteaming2024,
title={WildTeaming at Scale: From In-the-Wild Jailbreaks to (Adversarially) Safer Language Models},
author={Liwei Jiang and Kavel Rao and Seungju Han and Allyson Ettinger and Faeze Brahman and Sachin Kumar and Niloofar Mireshghallah and Ximing Lu and Maarten Sap and Yejin Choi and Nouha Dziri},
booktitle={The Thirty-eighth Annual Conference on Neural Information Processing Systems},
year={2024},
url={https://openreview.net/forum?id=n5R6TvBVcX}
}

@inproceedings{
brahman-kumar2024,
title={The Art of Saying No: Contextual Noncompliance in Language Models},
author={Faeze Brahman and Sachin Kumar and Vidhisha Balachandran and Pradeep Dasigi and Valentina Pyatkin and Abhilasha Ravichander and Sarah Wiegreffe and Nouha Dziri and Khyathi Chandu and Jack Hessel and Yulia Tsvetkov and Noah A. Smith and Yejin Choi and Hannaneh Hajishirzi},
booktitle={The Thirty-eight Conference on Neural Information Processing Systems Datasets and Benchmarks Track},
year={2024},
url={https://openreview.net/forum?id=f1UL4wNlw6}
}

@inproceedings{
luo2024wizardcoder,
title={WizardCoder: Empowering Code Large Language Models with Evol-Instruct},
author={Ziyang Luo and Can Xu and Pu Zhao and Qingfeng Sun and Xiubo Geng and Wenxiang Hu and Chongyang Tao and Jing Ma and Qingwei Lin and Daxin Jiang},
booktitle={The Twelfth International Conference on Learning Representations},
year={2024},
url={https://openreview.net/forum?id=UnUwSIgK5W}
}

@misc{numina_math_7b,
  author = {Edward Beeching and Shengyi Costa Huang and Albert Jiang and Jia Li and Benjamin Lipkin and Zihan Qina and Kashif Rasul and Ziju Shen and Roman Soletskyi and Lewis Tunstall},
  title = {NuminaMath 7B TIR},
  year = {2024},
  publisher = {Numina & Hugging Face},
  journal = {Hugging Face repository},
  howpublished = {\url{https://huggingface.co/AI-MO/NuminaMath-7B-TIR}}
}

@inproceedings{
toshniwal2025openmathinstruct,
title={OpenMathInstruct-2: Accelerating {AI} for Math with Massive Open-Source Instruction Data},
author={Shubham Toshniwal and Wei Du and Ivan Moshkov and Branislav Kisacanin and Alexan Ayrapetyan and Igor Gitman},
booktitle={The Thirteenth International Conference on Learning Representations},
year={2025},
url={https://openreview.net/forum?id=mTCbq2QssD}
}

@inproceedings{wadden-etal-2025-sciriff,
    title = "{S}ci{RIFF}: A Resource to Enhance Language Model Instruction-Following over Scientific Literature",
    author = "Wadden, David  and
      Shi, Kejian  and
      Morrison, Jacob  and
      Li, Alan  and
      Naik, Aakanksha  and
      Singh, Shruti  and
      Barzilay, Nitzan  and
      Lo, Kyle  and
      Hope, Tom  and
      Soldaini, Luca  and
      Shen, Shannon Zejiang  and
      Downey, Doug  and
      Hajishirzi, Hannaneh  and
      Cohan, Arman",
    editor = "Christodoulopoulos, Christos  and
      Chakraborty, Tanmoy  and
      Rose, Carolyn  and
      Peng, Violet",
    booktitle = "Proceedings of the 2025 Conference on Empirical Methods in Natural Language Processing",
    month = nov,
    year = "2025",
    address = "Suzhou, China",
    publisher = "Association for Computational Linguistics",
    url = "https://aclanthology.org/2025.emnlp-main.310/",
    doi = "10.18653/v1/2025.emnlp-main.310",
    pages = "6072--6109",
    ISBN = "979-8-89176-332-6"
}

@inproceedings{
weifinetuned,
title={Finetuned Language Models are Zero-Shot Learners},
author={Jason Wei and Maarten Bosma and Vincent Zhao and Kelvin Guu and Adams Wei Yu and Brian Lester and Nan Du and Andrew M. Dai and Quoc V Le},
booktitle={International Conference on Learning Representations},
year={2022},
url={https://openreview.net/forum?id=gEZrGCozdqR}
}

@inproceedings{deng2024wildvisopensourcevisualizer,
    title = "{W}ild{V}is: Open Source Visualizer for Million-Scale Chat Logs in the Wild",
    author = "Deng, Yuntian  and
      Zhao, Wenting  and
      Hessel, Jack  and
      Ren, Xiang  and
      Cardie, Claire  and
      Choi, Yejin",
    editor = "Hernandez Farias, Delia Irazu  and
      Hope, Tom  and
      Li, Manling",
    booktitle = "Proceedings of the 2024 Conference on Empirical Methods in Natural Language Processing: System Demonstrations",
    month = nov,
    year = "2024",
    address = "Miami, Florida, USA",
    publisher = "Association for Computational Linguistics",
    url = "https://aclanthology.org/2024.emnlp-demo.50/",
    doi = "10.18653/v1/2024.emnlp-demo.50",
    pages = "497--506"
}

@inproceedings{
  zhao2024wildchat,
  title={WildChat: 1M Chat{GPT} Interaction Logs in the Wild},
  author={Wenting Zhao and Xiang Ren and Jack Hessel and Claire Cardie and Yejin Choi and Yuntian Deng},
  booktitle={The Twelfth International Conference on Learning Representations},
  year={2024},
  url={https://openreview.net/forum?id=Bl8u7ZRlbM}
}

@inproceedings{
pf2023openassistant,
title={OpenAssistant Conversations - Democratizing Large Language Model Alignment},
author={Andreas K{\"o}pf and Yannic Kilcher and Dimitri von R{\"u}tte and Sotiris Anagnostidis and Zhi Rui Tam and Keith Stevens and Abdullah Barhoum and Duc Minh Nguyen and Oliver Stanley and Rich{\'a}rd Nagyfi and Shahul ES and Sameer Suri and David Alexandrovich Glushkov and Arnav Varma Dantuluri and Andrew Maguire and Christoph Schuhmann and Huu Nguyen and Alexander Julian Mattick},
booktitle={Thirty-seventh Conference on Neural Information Processing Systems Datasets and Benchmarks Track},
year={2023},
url={https://openreview.net/forum?id=VSJotgbPHF}
}

@misc{no_robots,
  author = {Nazneen Rajani and Lewis Tunstall and Edward Beeching and Nathan Lambert and Alexander M. Rush and Thomas Wolf},
  title = {No Robots},
  year = {2023},
  publisher = {Hugging Face},
  journal = {Hugging Face repository},
  howpublished = {\url{https://huggingface.co/datasets/HuggingFaceH4/no_robots}}
}

@inproceedings{singh2024aya,
    title = "Aya Dataset: An Open-Access Collection for Multilingual Instruction Tuning",
    author = {Singh, Shivalika  and
      Vargus, Freddie  and
      D{'}souza, Daniel  and
      Karlsson, B{\"o}rje F.  and
      Mahendiran, Abinaya  and
      Ko, Wei-Yin  and
      Shandilya, Herumb  and
      Patel, Jay  and
      Mataciunas, Deividas  and
      O{'}Mahony, Laura  and
      Zhang, Mike  and
      Hettiarachchi, Ramith  and
      Wilson, Joseph  and
      Machado, Marina  and
      Moura, Luisa  and
      Krzemi{\'n}ski, Dominik  and
      Fadaei, Hakimeh  and
      Ergun, Irem  and
      Okoh, Ifeoma  and
      Alaagib, Aisha  and
      Mudannayake, Oshan  and
      Alyafeai, Zaid  and
      Chien, Vu  and
      Ruder, Sebastian  and
      Guthikonda, Surya  and
      Alghamdi, Emad  and
      Gehrmann, Sebastian  and
      Muennighoff, Niklas  and
      Bartolo, Max  and
      Kreutzer, Julia  and
      {\"U}st{\"u}n, Ahmet  and
      Fadaee, Marzieh  and
      Hooker, Sara},
    editor = "Ku, Lun-Wei  and
      Martins, Andre  and
      Srikumar, Vivek",
    booktitle = "Proceedings of the 62nd Annual Meeting of the Association for Computational Linguistics (Volume 1: Long Papers)",
    month = aug,
    year = "2024",
    address = "Bangkok, Thailand",
    publisher = "Association for Computational Linguistics",
    url = "https://aclanthology.org/2024.acl-long.620/",
    doi = "10.18653/v1/2024.acl-long.620",
    pages = "11521--11567"
}

@misc{tao2026modalmix,
  title={{ModalMix}: Optimizing Multimodal Data Mixture with Compute-Dependent Regression},
  author={{Anonymous}},
  year={2026},
  howpublished={OpenReview preprint},
  note={Anonymous submission to ICLR 2026},
  url={https://openreview.net/pdf?id=R1HcIN90A1}
}

@inproceedings{wang2025tikmixdatainfluencedynamic,
  title={{TiKMiX}: Efficient Semi-Dynamic Data Mixture via Data Influence for {LLM} Pre-training},
  author={Yifan Wang and Binbin Liu and Fengze Liu and Yuanfan Guo and Jiyao Deng and Xuecheng Wu and Weidong Zhou and Xiaohuan Zhou and Taifeng Wang},
  booktitle={Proceedings of the 64th Annual Meeting of the Association for Computational Linguistics (Volume 1: Long Papers)},
  pages={5777--5793},
  publisher={Association for Computational Linguistics},
  year={2026},
  url={https://aclanthology.org/2026.acl-long.261/}
}

@misc{
xu2026unveiling,
title={Unveiling Downstream Performance Scaling of {LLM}s: A Clustering-Based Perspective},
author={Chengyin Xu and Kaiyuan Chen and Xiao Li and Ke Shen and Chenggang Li},
year={2025},
eprint={2502.17262},
archivePrefix={arXiv},
primaryClass={cs.CL},
url={https://arxiv.org/abs/2502.17262}
}

@inproceedings{gao-etal-2025-strategic,
    title = "A Strategic Coordination Framework of Small {LM}s Matches Large {LM}s in Data Synthesis",
    author = "Gao, Xin  and
      Pei, Qizhi  and
      Tang, Zinan  and
      Li, Yu  and
      Lin, Honglin  and
      Wu, Jiang  and
      Wu, Lijun  and
      He, Conghui",
    editor = "Che, Wanxiang  and
      Nabende, Joyce  and
      Shutova, Ekaterina  and
      Pilehvar, Mohammad Taher",
    booktitle = "Proceedings of the 63rd Annual Meeting of the Association for Computational Linguistics (Volume 1: Long Papers)",
    month = jul,
    year = "2025",
    address = "Vienna, Austria",
    publisher = "Association for Computational Linguistics",
    url = "https://aclanthology.org/2025.acl-long.566/",
    doi = "10.18653/v1/2025.acl-long.566",
    pages = "11552--11570",
    ISBN = "979-8-89176-251-0"
}

@inproceedings{tang-etal-2025-middo,
    title = "Middo: Model-Informed Dynamic Data Optimization for Enhanced {LLM} Fine-Tuning via Closed-Loop Learning",
    author = "Tang, Zinan  and
      Gao, Xin  and
      Pei, Qizhi  and
      Pan, Zhuoshi  and
      Cai, Mengzhang  and
      Wu, Jiang  and
      He, Conghui  and
      Wu, Lijun",
    editor = "Christodoulopoulos, Christos  and
      Chakraborty, Tanmoy  and
      Rose, Carolyn  and
      Peng, Violet",
    booktitle = "Proceedings of the 2025 Conference on Empirical Methods in Natural Language Processing",
    month = nov,
    year = "2025",
    address = "Suzhou, China",
    publisher = "Association for Computational Linguistics",
    url = "https://aclanthology.org/2025.emnlp-main.350/",
    doi = "10.18653/v1/2025.emnlp-main.350",
    pages = "6871--6891",
    ISBN = "979-8-89176-332-6"
}

@inproceedings{
kandpal2025position,
title={Position: The Most Expensive Part of an {LLM} *should* be its Training Data},
author={Nikhil Kandpal and Colin Raffel},
booktitle={Forty-second International Conference on Machine Learning Position Paper Track},
year={2025},
url={https://openreview.net/forum?id=L6RpQ1h4Nx}
}

@article{zhang2025survey,
  title={A survey on data selection for llm instruction tuning},
  author={Zhang, Bolin and Wang, Jiahao and Du, Qianlong and Zhang, Jiajun and Tu, Zhiying and Chu, Dianhui},
  journal={Journal of Artificial Intelligence Research},
  volume={83},
  year={2025}
}

@misc{balappanawar2025if,
  title={Flying Pigs, {FaR} and Beyond: Evaluating {LLM} Reasoning in Counterfactual Worlds},
  author={Joishy, Anish R and Balappanawar, Ishwar B and Bonagiri, Vamshi Krishna and Gaur, Manas and Thirunarayan, Krishnaprasad and Kumaraguru, Ponnurangam},
  year={2026},
  eprint={2505.22318v2},
  archivePrefix={arXiv},
  primaryClass={cs.CL},
  url={https://arxiv.org/abs/2505.22318v2}
}

@article{wu2025knowledge,
  title={Knowledge or reasoning? a close look at how llms think across domains},
  author={Wu, Juncheng and Liu, Sheng and Tu, Haoqin and Yu, Hang and Huang, Xiaoke and Zou, James and Xie, Cihang and Zhou, Yuyin},
  journal={arXiv preprint arXiv:2506.02126},
  year={2025}
}

@article{robinson1988root,
  title={Root-N-consistent semiparametric regression},
  author={Robinson, Peter M},
  journal={Econometrica: journal of the Econometric Society},
  pages={931--954},
  year={1988},
  publisher={JSTOR}
}

@misc{qwen3technicalreport,
      title={Qwen3 Technical Report}, 
      author={{Qwen Team}},
      year={2025},
      eprint={2505.09388},
      archivePrefix={arXiv},
      primaryClass={cs.CL},
      url={https://arxiv.org/abs/2505.09388}, 
}

@article{tian2025correctanswersequaldistillation,
      title={Not All Correct Answers Are Equal: Why Your Distillation Source Matters}, 
      author={Xiaoyu Tian and Yunjie Ji and Haotian Wang and Shuaiting Chen and Sitong Zhao and Yiping Peng and Han Zhao and Xiangang Li},
      year={2025},
      journal={arXiv preprint arXiv:2505.14464},
      doi={10.48550/arXiv.2505.14464},
      eprint={2505.14464},
      archivePrefix={arXiv},
      primaryClass={cs.CL},
      url={https://arxiv.org/abs/2505.14464}, 
}

@misc{albalak2023efficientonlinedatamixing,
      title={Efficient Online Data Mixing For Language Model Pre-Training}, 
      author={Alon Albalak and Liangming Pan and Colin Raffel and William Yang Wang},
      year={2023},
      eprint={2312.02406},
      archivePrefix={arXiv},
      primaryClass={cs.CL},
      url={https://arxiv.org/abs/2312.02406}, 
}

@inproceedings{oprescu2019orthogonal,
  title={Orthogonal random forest for causal inference},
  author={Oprescu, Miruna and Syrgkanis, Vasilis and Wu, Zhiwei Steven},
  booktitle={International Conference on Machine Learning},
  pages={4932--4941},
  year={2019},
  organization={PMLR}
}

@article{wager2018estimation,
  title={Estimation and inference of heterogeneous treatment effects using random forests},
  author={Wager, Stefan and Athey, Susan},
  journal={Journal of the American Statistical Association},
  volume={113},
  number={523},
  pages={1228--1242},
  year={2018},
  publisher={Taylor \& Francis}
}

@inproceedings{10.5555/3294996.3295074,
author = {Ke, Guolin and Meng, Qi and Finley, Thomas and Wang, Taifeng and Chen, Wei and Ma, Weidong and Ye, Qiwei and Liu, Tie-Yan},
title = {LightGBM: a highly efficient gradient boosting decision tree},
year = {2017},
isbn = {9781510860964},
publisher = {Curran Associates Inc.},
address = {Red Hook, NY, USA},
booktitle = {Proceedings of the 31st International Conference on Neural Information Processing Systems},
pages = {3149--3157},
numpages = {9},
location = {Long Beach, California, USA},
series = {NIPS'17}
}

@misc{econml,
  author={Keith Battocchi and Eleanor Dillon and Maggie Hei and Greg Lewis and Paul Oka and Miruna Oprescu and Vasilis Syrgkanis},
  title={{EconML}: {A Python Package for ML-Based Heterogeneous Treatment Effects Estimation}},
  url={https://github.com/py-why/EconML},
  year={2019}
}

@inproceedings{lin-etal-2022-truthfulqa,
    title = "{T}ruthful{QA}: Measuring How Models Mimic Human Falsehoods",
    author = "Lin, Stephanie  and
      Hilton, Jacob  and
      Evans, Owain",
    editor = "Muresan, Smaranda  and
      Nakov, Preslav  and
      Villavicencio, Aline",
    booktitle = "Proceedings of the 60th Annual Meeting of the Association for Computational Linguistics (Volume 1: Long Papers)",
    month = may,
    year = "2022",
    address = "Dublin, Ireland",
    publisher = "Association for Computational Linguistics",
    url = "https://aclanthology.org/2022.acl-long.229/",
    doi = "10.18653/v1/2022.acl-long.229",
    pages = "3214--3252"
}

@inproceedings{hartvigsen-etal-2022-toxigen,
    title = "{T}oxi{G}en: A Large-Scale Machine-Generated Dataset for Adversarial and Implicit Hate Speech Detection",
    author = "Hartvigsen, Thomas  and
      Gabriel, Saadia  and
      Palangi, Hamid  and
      Sap, Maarten  and
      Ray, Dipankar  and
      Kamar, Ece",
    editor = "Muresan, Smaranda  and
      Nakov, Preslav  and
      Villavicencio, Aline",
    booktitle = "Proceedings of the 60th Annual Meeting of the Association for Computational Linguistics (Volume 1: Long Papers)",
    month = may,
    year = "2022",
    address = "Dublin, Ireland",
    publisher = "Association for Computational Linguistics",
    url = "https://aclanthology.org/2022.acl-long.234/",
    doi = "10.18653/v1/2022.acl-long.234",
    pages = "3309--3326"
}

@inproceedings{
pyatkin2026generalizing,
title={Generalizing Verifiable Instruction Following},
author={Valentina Pyatkin and Saumya Malik and Victoria Graf and Hamish Ivison and Shengyi Huang and Pradeep Dasigi and Nathan Lambert and Hannaneh Hajishirzi},
booktitle={The Thirty-ninth Annual Conference on Neural Information Processing Systems Datasets and Benchmarks Track},
year={2025},
url={https://openreview.net/forum?id=yfYgwjj5F8}
}

@article{zhou2023instruction,
  title={Instruction-following evaluation for large language models},
  author={Zhou, Jeffrey and Lu, Tianjian and Mishra, Swaroop and Brahma, Siddhartha and Basu, Sujoy and Luan, Yi and Zhou, Denny and Hou, Le},
  journal={arXiv preprint arXiv:2311.07911},
  year={2023}
}

@misc{austin2021programsynthesislargelanguage,
      title={Program Synthesis with Large Language Models}, 
      author={Jacob Austin and Augustus Odena and Maxwell Nye and Maarten Bosma and Henryk Michalewski and David Dohan and Ellen Jiang and Carrie Cai and Michael Terry and Quoc Le and Charles Sutton},
      year={2021},
      eprint={2108.07732},
      archivePrefix={arXiv},
      primaryClass={cs.PL},
      url={https://arxiv.org/abs/2108.07732}, 
}

@inproceedings{
liu2023is,
title={Is Your Code Generated by Chat{GPT} Really Correct? Rigorous Evaluation of Large Language Models for Code Generation},
author={Jiawei Liu and Chunqiu Steven Xia and Yuyao Wang and LINGMING ZHANG},
booktitle={Thirty-seventh Conference on Neural Information Processing Systems},
year={2023},
url={https://openreview.net/forum?id=1qvx610Cu7}
}

@misc{chen2021evaluatinglargelanguagemodels,
      title={Evaluating Large Language Models Trained on Code}, 
      author={Mark Chen and Jerry Tworek and Heewoo Jun and Qiming Yuan and Henrique Ponde de Oliveira Pinto and Jared Kaplan and Harri Edwards and Yuri Burda and Nicholas Joseph and Greg Brockman and Alex Ray and Raul Puri and Gretchen Krueger and Michael Petrov and Heidy Khlaaf and Girish Sastry and Pamela Mishkin and Brooke Chan and Scott Gray and Nick Ryder and Mikhail Pavlov and Alethea Power and Lukasz Kaiser and Mohammad Bavarian and Clemens Winter and Philippe Tillet and Felipe Petroski Such and Dave Cummings and Matthias Plappert and Fotios Chantzis and Elizabeth Barnes and Ariel Herbert-Voss and William Hebgen Guss and Alex Nichol and Alex Paino and Nikolas Tezak and Jie Tang and Igor Babuschkin and Suchir Balaji and Shantanu Jain and William Saunders and Christopher Hesse and Andrew N. Carr and Jan Leike and Josh Achiam and Vedant Misra and Evan Morikawa and Alec Radford and Matthew Knight and Miles Brundage and Mira Murati and Katie Mayer and Peter Welinder and Bob McGrew and Dario Amodei and Sam McCandlish and Ilya Sutskever and Wojciech Zaremba},
      year={2021},
      eprint={2107.03374},
      archivePrefix={arXiv},
      primaryClass={cs.LG},
      url={https://arxiv.org/abs/2107.03374}, 
}

@inproceedings{
math,
title={Measuring Mathematical Problem Solving With the {MATH} Dataset},
author={Dan Hendrycks and Collin Burns and Saurav Kadavath and Akul Arora and Steven Basart and Eric Tang and Dawn Song and Jacob Steinhardt},
booktitle={Thirty-fifth Conference on Neural Information Processing Systems Datasets and Benchmarks Track (Round 2)},
year={2021},
url={https://openreview.net/forum?id=7Bywt2mQsCe}
}

@inproceedings{he-etal-2024-olympiadbench,
    title = "{O}lympiad{B}ench: A Challenging Benchmark for Promoting {AGI} with Olympiad-Level Bilingual Multimodal Scientific Problems",
    author = "He, Chaoqun  and
      Luo, Renjie  and
      Bai, Yuzhuo  and
      Hu, Shengding  and
      Thai, Zhen  and
      Shen, Junhao  and
      Hu, Jinyi  and
      Han, Xu  and
      Huang, Yujie  and
      Zhang, Yuxiang  and
      Liu, Jie  and
      Qi, Lei  and
      Liu, Zhiyuan  and
      Sun, Maosong",
    editor = "Ku, Lun-Wei  and
      Martins, Andre  and
      Srikumar, Vivek",
    booktitle = "Proceedings of the 62nd Annual Meeting of the Association for Computational Linguistics (Volume 1: Long Papers)",
    month = aug,
    year = "2024",
    address = "Bangkok, Thailand",
    publisher = "Association for Computational Linguistics",
    url = "https://aclanthology.org/2024.acl-long.211/",
    doi = "10.18653/v1/2024.acl-long.211",
    pages = "3828--3850"
}

@inproceedings{
hendrycks2021measuring,
title={Measuring Massive Multitask Language Understanding},
author={Dan Hendrycks and Collin Burns and Steven Basart and Andy Zou and Mantas Mazeika and Dawn Song and Jacob Steinhardt},
booktitle={International Conference on Learning Representations},
year={2021},
url={https://openreview.net/forum?id=d7KBjmI3GmQ}
}

@misc{cobbe2021trainingverifierssolvemath,
      title={Training Verifiers to Solve Math Word Problems}, 
      author={Karl Cobbe and Vineet Kosaraju and Mohammad Bavarian and Mark Chen and Heewoo Jun and Lukasz Kaiser and Matthias Plappert and Jerry Tworek and Jacob Hilton and Reiichiro Nakano and Christopher Hesse and John Schulman},
      year={2021},
      eprint={2110.14168},
      archivePrefix={arXiv},
      primaryClass={cs.LG},
      url={https://arxiv.org/abs/2110.14168}, 
}

@inproceedings{zhong-etal-2024-agieval,
    title = "{AGIE}val: A Human-Centric Benchmark for Evaluating Foundation Models",
    author = "Zhong, Wanjun  and
      Cui, Ruixiang  and
      Guo, Yiduo  and
      Liang, Yaobo  and
      Lu, Shuai  and
      Wang, Yanlin  and
      Saied, Amin  and
      Chen, Weizhu  and
      Duan, Nan",
    editor = "Duh, Kevin  and
      Gomez, Helena  and
      Bethard, Steven",
    booktitle = "Findings of the Association for Computational Linguistics: NAACL 2024",
    month = jun,
    year = "2024",
    address = "Mexico City, Mexico",
    publisher = "Association for Computational Linguistics",
    url = "https://aclanthology.org/2024.findings-naacl.149/",
    doi = "10.18653/v1/2024.findings-naacl.149",
    pages = "2299--2314"
}

@inproceedings{dua-etal-2019-drop,
    title = "{DROP}: A Reading Comprehension Benchmark Requiring Discrete Reasoning Over Paragraphs",
    author = "Dua, Dheeru  and
      Wang, Yizhong  and
      Dasigi, Pradeep  and
      Stanovsky, Gabriel  and
      Singh, Sameer  and
      Gardner, Matt",
    editor = "Burstein, Jill  and
      Doran, Christy  and
      Solorio, Thamar",
    booktitle = "Proceedings of the 2019 Conference of the North {A}merican Chapter of the Association for Computational Linguistics: Human Language Technologies, Volume 1 (Long and Short Papers)",
    month = jun,
    year = "2019",
    address = "Minneapolis, Minnesota",
    publisher = "Association for Computational Linguistics",
    url = "https://aclanthology.org/N19-1246/",
    doi = "10.18653/v1/N19-1246",
    pages = "2368--2378"
}

@article{
srivastava2023beyond,
title={Beyond the Imitation Game: Quantifying and extrapolating the capabilities of language models},
author={Aarohi Srivastava and Abhinav Rastogi and Abhishek Rao and Abu Awal Md Shoeb and Abubakar Abid and Adam Fisch and Adam R. Brown and Adam Santoro and Aditya Gupta and Adri{\`a} Garriga-Alonso and Agnieszka Kluska and Aitor Lewkowycz and Akshat Agarwal and Alethea Power and Alex Ray and Alex Warstadt and Alexander W. Kocurek and Ali Safaya and Ali Tazarv and Alice Xiang and Alicia Parrish and Allen Nie and Aman Hussain and Amanda Askell and Amanda Dsouza and Ambrose Slone and Ameet Rahane and Anantharaman S. Iyer and Anders Johan Andreassen and Andrea Madotto and Andrea Santilli and Andreas Stuhlm{\"u}ller and Andrew M. Dai and Andrew La and Andrew Kyle Lampinen and Andy Zou and Angela Jiang and Angelica Chen and Anh Vuong and Animesh Gupta and Anna Gottardi and Antonio Norelli and Anu Venkatesh and Arash Gholamidavoodi and Arfa Tabassum and Arul Menezes and Arun Kirubarajan and Asher Mullokandov and Ashish Sabharwal and Austin Herrick and Avia Efrat and Aykut Erdem and Ayla Karaka{\c{s}} and B. Ryan Roberts and Bao Sheng Loe and Barret Zoph and Bart{\l}omiej Bojanowski and Batuhan {\"O}zyurt and Behnam Hedayatnia and Behnam Neyshabur and Benjamin Inden and Benno Stein and Berk Ekmekci and Bill Yuchen Lin and Blake Howald and Bryan Orinion and Cameron Diao and Cameron Dour and Catherine Stinson and Cedrick Argueta and Cesar Ferri and Chandan Singh and Charles Rathkopf and Chenlin Meng and Chitta Baral and Chiyu Wu and Chris Callison-Burch and Christopher Waites and Christian Voigt and Christopher D Manning and Christopher Potts and Cindy Ramirez and Clara E. Rivera and Clemencia Siro and Colin Raffel and Courtney Ashcraft and Cristina Garbacea and Damien Sileo and Dan Garrette and Dan Hendrycks and Dan Kilman and Dan Roth and C. Daniel Freeman and Daniel Khashabi and Daniel Levy and Daniel Mosegu{\'\i} Gonz{\'a}lez and Danielle Perszyk and Danny Hernandez and Danqi Chen and Daphne Ippolito and Dar Gilboa and David Dohan and David Drakard and David Jurgens and Debajyoti Datta and Deep Ganguli and Denis Emelin and Denis Kleyko and Deniz Yuret and Derek Chen and Derek Tam and Dieuwke Hupkes and Diganta Misra and Dilyar Buzan and Dimitri Coelho Mollo and Diyi Yang and Dong-Ho Lee and Dylan Schrader and Ekaterina Shutova and Ekin Dogus Cubuk and Elad Segal and Eleanor Hagerman and Elizabeth Barnes and Elizabeth Donoway and Ellie Pavlick and Emanuele Rodol{\`a} and Emma Lam and Eric Chu and Eric Tang and Erkut Erdem and Ernie Chang and Ethan A Chi and Ethan Dyer and Ethan Jerzak and Ethan Kim and Eunice Engefu Manyasi and Evgenii Zheltonozhskii and Fanyue Xia and Fatemeh Siar and Fernando Mart{\'\i}nez-Plumed and Francesca Happ{\'e} and Francois Chollet and Frieda Rong and Gaurav Mishra and Genta Indra Winata and Gerard de Melo and Germ{\`a}n Kruszewski and Giambattista Parascandolo and Giorgio Mariani and Gloria Xinyue Wang and Gonzalo Jaimovitch-Lopez and Gregor Betz and Guy Gur-Ari and Hana Galijasevic and Hannah Kim and Hannah Rashkin and Hannaneh Hajishirzi and Harsh Mehta and Hayden Bogar and Henry Francis Anthony Shevlin and Hinrich Schuetze and Hiromu Yakura and Hongming Zhang and Hugh Mee Wong and Ian Ng and Isaac Noble and Jaap Jumelet and Jack Geissinger and Jackson Kernion and Jacob Hilton and Jaehoon Lee and Jaime Fern{\'a}ndez Fisac and James B Simon and James Koppel and James Zheng and James Zou and Jan Kocon and Jana Thompson and Janelle Wingfield and Jared Kaplan and Jarema Radom and Jascha Sohl-Dickstein and Jason Phang and Jason Wei and Jason Yosinski and Jekaterina Novikova and Jelle Bosscher and Jennifer Marsh and Jeremy Kim and Jeroen Taal and Jesse Engel and Jesujoba Alabi and Jiacheng Xu and Jiaming Song and Jillian Tang and Joan Waweru and John Burden and John Miller and John U. Balis and Jonathan Batchelder and Jonathan Berant and J{\"o}rg Frohberg and Jos Rozen and Jose Hernandez-Orallo and Joseph Boudeman and Joseph Guerr and Joseph Jones and Joshua B. Tenenbaum and Joshua S. Rule and Joyce Chua and Kamil Kanclerz and Karen Livescu and Karl Krauth and Karthik Gopalakrishnan and Katerina Ignatyeva and Katja Markert and Kaustubh Dhole and Kevin Gimpel and Kevin Omondi and Kory Wallace Mathewson and Kristen Chiafullo and Ksenia Shkaruta and Kumar Shridhar and Kyle McDonell and Kyle Richardson and Laria Reynolds and Leo Gao and Li Zhang and Liam Dugan and Lianhui Qin and Lidia Contreras-Ochando and Louis-Philippe Morency and Luca Moschella and Lucas Lam and Lucy Noble and Ludwig Schmidt and Luheng He and Luis Oliveros-Col{\'o}n and Luke Metz and L{\"u}tfi Kerem Senel and Maarten Bosma and Maarten Sap and Maartje Ter Hoeve and Maheen Farooqi and Manaal Faruqui and Mantas Mazeika and Marco Baturan and Marco Marelli and Marco Maru and Maria Jose Ramirez-Quintana and Marie Tolkiehn and Mario Giulianelli and Martha Lewis and Martin Potthast and Matthew L Leavitt and Matthias Hagen and M{\'a}ty{\'a}s Schubert and Medina Orduna Baitemirova and Melody Arnaud and Melvin McElrath and Michael Andrew Yee and Michael Cohen and Michael Gu and Michael Ivanitskiy and Michael Starritt and Michael Strube and Micha{\l} Sw{\k{e}}drowski and Michele Bevilacqua and Michihiro Yasunaga and Mihir Kale and Mike Cain and Mimee Xu and Mirac Suzgun and Mitch Walker and Mo Tiwari and Mohit Bansal and Moin Aminnaseri and Mor Geva and Mozhdeh Gheini and Mukund Varma T and Nanyun Peng and Nathan Andrew Chi and Nayeon Lee and Neta Gur-Ari Krakover and Nicholas Cameron and Nicholas Roberts and Nick Doiron and Nicole Martinez and Nikita Nangia and Niklas Deckers and Niklas Muennighoff and Nitish Shirish Keskar and Niveditha S. Iyer and Noah Constant and Noah Fiedel and Nuan Wen and Oliver Zhang and Omar Agha and Omar Elbaghdadi and Omer Levy and Owain Evans and Pablo Antonio Moreno Casares and Parth Doshi and Pascale Fung and Paul Pu Liang and Paul Vicol and Pegah Alipoormolabashi and Peiyuan Liao and Percy Liang and Peter W Chang and Peter Eckersley and Phu Mon Htut and Pinyu Hwang and Piotr Mi{\l}kowski and Piyush Patil and Pouya Pezeshkpour and Priti Oli and Qiaozhu Mei and Qing Lyu and Qinlang Chen and Rabin Banjade and Rachel Etta Rudolph and Raefer Gabriel and Rahel Habacker and Ramon Risco and Rapha{\"e}l Milli{\`e}re and Rhythm Garg and Richard Barnes and Rif A. Saurous and Riku Arakawa and Robbe Raymaekers and Robert Frank and Rohan Sikand and Roman Novak and Roman Sitelew and Ronan Le Bras and Rosanne Liu and Rowan Jacobs and Rui Zhang and Russ Salakhutdinov and Ryan Andrew Chi and Seungjae Ryan Lee and Ryan Stovall and Ryan Teehan and Rylan Yang and Sahib Singh and Saif M. Mohammad and Sajant Anand and Sam Dillavou and Sam Shleifer and Sam Wiseman and Samuel Gruetter and Samuel R. Bowman and Samuel Stern Schoenholz and Sanghyun Han and Sanjeev Kwatra and Sarah A. Rous and Sarik Ghazarian and Sayan Ghosh and Sean Casey and Sebastian Bischoff and Sebastian Gehrmann and Sebastian Schuster and Sepideh Sadeghi and Shadi Hamdan and Sharon Zhou and Shashank Srivastava and Sherry Shi and Shikhar Singh and Shima Asaadi and Shixiang Shane Gu and Shubh Pachchigar and Shubham Toshniwal and Shyam Upadhyay and Shyamolima Shammie Debnath and Siamak Shakeri and Simon Thormeyer and Simone Melzi and Siva Reddy and Sneha Priscilla Makini and Soo-Hwan Lee and Spencer Torene and Sriharsha Hatwar and Stanislas Dehaene and Stefan Divic and Stefano Ermon and Stella Biderman and Stephanie Lin and Stephen Prasad and Steven Piantadosi and Stuart Shieber and Summer Misherghi and Svetlana Kiritchenko and Swaroop Mishra and Tal Linzen and Tal Schuster and Tao Li and Tao Yu and Tariq Ali and Tatsunori Hashimoto and Te-Lin Wu and Th{\'e}o Desbordes and Theodore Rothschild and Thomas Phan and Tianle Wang and Tiberius Nkinyili and Timo Schick and Timofei Kornev and Titus Tunduny and Tobias Gerstenberg and Trenton Chang and Trishala Neeraj and Tushar Khot and Tyler Shultz and Uri Shaham and Vedant Misra and Vera Demberg and Victoria Nyamai and Vikas Raunak and Vinay Venkatesh Ramasesh and vinay uday prabhu and Vishakh Padmakumar and Vivek Srikumar and William Fedus and William Saunders and William Zhang and Wout Vossen and Xiang Ren and Xiaoyu Tong and Xinran Zhao and Xinyi Wu and Xudong Shen and Yadollah Yaghoobzadeh and Yair Lakretz and Yangqiu Song and Yasaman Bahri and Yejin Choi and Yichi Yang and Sophie Hao and Yifu Chen and Yonatan Belinkov and Yu Hou and Yufang Hou and Yuntao Bai and Zachary Seid and Zhuoye Zhao and Zijian Wang and Zijie J. Wang and Zirui Wang and Ziyi Wu},
journal={Transactions on Machine Learning Research},
issn={2835-8856},
year={2023},
url={https://openreview.net/forum?id=uyTL5Bvosj},
note={Featured Certification}
}

@inproceedings{
rein2024gpqa,
title={{GPQA}: A Graduate-Level Google-Proof Q\&A Benchmark},
author={David Rein and Betty Li Hou and Asa Cooper Stickland and Jackson Petty and Richard Yuanzhe Pang and Julien Dirani and Julian Michael and Samuel R. Bowman},
booktitle={First Conference on Language Modeling},
year={2024},
url={https://openreview.net/forum?id=Ti67584b98}
}

@inproceedings{
wang2024mmlupro,
title={{MMLU}-Pro: A More Robust and Challenging Multi-Task Language Understanding Benchmark},
author={Yubo Wang and Xueguang Ma and Ge Zhang and Yuansheng Ni and Abhranil Chandra and Shiguang Guo and Weiming Ren and Aaran Arulraj and Xuan He and Ziyan Jiang and Tianle Li and Max Ku and Kai Wang and Alex Zhuang and Rongqi Fan and Xiang Yue and Wenhu Chen},
booktitle={The Thirty-eight Conference on Neural Information Processing Systems Datasets and Benchmarks Track},
year={2024},
url={https://openreview.net/forum?id=y10DM6R2r3}
}

@inproceedings{
lambert2025tulu,
title={Tulu 3: Pushing Frontiers in Open Language Model Post-Training},
author={Nathan Lambert and Jacob Morrison and Valentina Pyatkin and Shengyi Huang and Hamish Ivison and Faeze Brahman and Lester James Validad Miranda and Alisa Liu and Nouha Dziri and Xinxi Lyu and Yuling Gu and Saumya Malik and Victoria Graf and Jena D. Hwang and Jiangjiang Yang and Ronan Le Bras and Oyvind Tafjord and Christopher Wilhelm and Luca Soldaini and Noah A. Smith and Yizhong Wang and Pradeep Dasigi and Hannaneh Hajishirzi},
booktitle={Second Conference on Language Modeling},
year={2025},
url={https://openreview.net/forum?id=i1uGbfHHpH}
}

@misc{
li2026unified,
title={Unified Data Selection for {LLM} Reasoning},
author={Xiaoyuan Li and Yubo Ma and Chengpeng Li and Fengbin Zhu and Yiyao Yu and Keqin Bao and Wenjie Wang and Fuli Feng and Dayiheng Liu},
year={2026},
eprint={2605.22389},
archivePrefix={arXiv},
primaryClass={cs.CL},
url={https://arxiv.org/abs/2605.22389}
}

@inproceedings{10.5555/3692070.3694241,
author = {Wettig, Alexander and Gupta, Aatmik and Malik, Saumya and Chen, Danqi},
title = {QuRating: selecting high-quality data for training language models},
year = {2024},
publisher = {JMLR.org},
booktitle = {Proceedings of the 41st International Conference on Machine Learning},
articleno = {2171},
numpages = {57},
location = {Vienna, Austria},
series = {ICML'24}
}

@inproceedings{shum2025predictive,
  title={Predictive Data Selection: The Data That Predicts Is the Data That Teaches},
  author={Shum, Kashun and Huang, Yuzhen and Zou, Hongjian and Ding, Qi and Liao, Yixuan and Chen, Xiaoxin and Liu, Qian and He, Junxian},
  booktitle={Proceedings of the 42nd International Conference on Machine Learning},
  volume={267},
  series={Proceedings of Machine Learning Research},
  pages={55427--55450},
  publisher={PMLR},
  year={2025},
  url={https://proceedings.mlr.press/v267/shum25a.html}
}

@misc{2023opencompass,
    title={OpenCompass: A Universal Evaluation Platform for Foundation Models},
    author={{OpenCompass Contributors}},
    howpublished = {\url{https://github.com/open-compass/opencompass}},
    year={2023}
}

@inproceedings{zheng2024llamafactory,
  title={{LlamaFactory}: Unified Efficient Fine-Tuning of 100+ Language Models},
  author={Yaowei Zheng and Richong Zhang and Junhao Zhang and Yanhan Ye and Zheyan Luo},
  booktitle={Proceedings of the 62nd Annual Meeting of the Association for Computational Linguistics (Volume 3: System Demonstrations)},
  address={Bangkok, Thailand},
  publisher={Association for Computational Linguistics},
  pages={400--410},
  year={2024},
  url={https://aclanthology.org/2024.acl-demos.38/}
}

@misc{qwen2.5,
      title={Qwen2.5 Technical Report}, 
      author={An Yang and Baosong Yang and Beichen Zhang and Binyuan Hui and Bo Zheng and Bowen Yu and Chengyuan Li and Dayiheng Liu and Fei Huang and Haoran Wei and Huan Lin and Jian Yang and Jianhong Tu and Jianwei Zhang and Jianxin Yang and Jiaxi Yang and Jingren Zhou and Junyang Lin and Kai Dang and Keming Lu and Keqin Bao and Kexin Yang and Le Yu and Mei Li and Mingfeng Xue and Pei Zhang and Qin Zhu and Rui Men and Runji Lin and Tianhao Li and Tianyi Tang and Tingyu Xia and Xingzhang Ren and Xuancheng Ren and Yang Fan and Yang Su and Yichang Zhang and Yu Wan and Yuqiong Liu and Zeyu Cui and Zhenru Zhang and Zihan Qiu},
      year={2024},
      eprint={2412.15115},
      archivePrefix={arXiv},
      primaryClass={cs.CL},
      url={https://arxiv.org/abs/2412.15115}, 
}

@article{qwen2,
      title={Qwen2 Technical Report}, 
      author={Yang, An and Yang, Baosong and Hui, Binyuan and Zheng, Bo and Yu, Bowen and Zhou, Chang and Li, Chengpeng and Li, Chengyuan and Liu, Dayiheng and Huang, Fei and Dong, Guanting and Wei, Haoran and Lin, Huan and Tang, Jialong and Wang, Jialin and Yang, Jian and Tu, Jianhong and Zhang, Jianwei and Ma, Jianxin and Yang, Jianxin and Xu, Jin and Zhou, Jingren and Bai, Jinze and He, Jinzheng and Lin, Junyang and Dang, Kai and Lu, Keming and Chen, Keqin and Yang, Kexin and Li, Mei and Xue, Mingfeng and Ni, Na and Zhang, Pei and Wang, Peng and Peng, Ru and Men, Rui and Gao, Ruize and Lin, Runji and Wang, Shijie and Bai, Shuai and Tan, Sinan and Zhu, Tianhang and Li, Tianhao and Liu, Tianyu and Ge, Wenbin and Deng, Xiaodong and Zhou, Xiaohuan and Ren, Xingzhang and Zhang, Xinyu and Wei, Xipin and Ren, Xuancheng and Liu, Xuejing and Fan, Yang and Yao, Yang and Zhang, Yichang and Wan, Yu and Chu, Yunfei and Liu, Yuqiong and Cui, Zeyu and Zhang, Zhenru and Guo, Zhifang and Fan, Zhihao},
      journal={arXiv preprint arXiv:2407.10671},
      year={2024}
}

@misc{opendataarena_tool_2025,
  author       = {OpenDataArena},
  title        = {{OpenDataArena-Tool}},
  year         = {2025},
  url          = {https://github.com/OpenDataArena/OpenDataArena-Tool},
  note         = {GitHub repository},
}

@article{cai2025opendataarena,
  title={OpenDataArena: A Fair and Open Arena for Benchmarking Post-Training Dataset Value},
  author={Cai, Mengzhang and Gao, Xin and Li, Yu and Lin, Honglin and Liu, Zheng and Pan, Zhuoshi and Pei, Qizhi and Shang, Xiaoran and Sun, Mengyuan and Tang, Zinan and others},
  journal={arXiv preprint arXiv:2512.14051},
  year={2025}
}

@misc{opendataarena_scored_data_2025,
  author       = {OpenDataArena},
  title        = {OpenDataArena-scored-data},
  year         = {2025},
  url          = {https://huggingface.co/datasets/OpenDataArena/OpenDataArena-scored-data}
}

@article{kaplan2020scaling,
  title={Scaling laws for neural language models},
  author={Kaplan, Jared and McCandlish, Sam and Henighan, Tom and Brown, Tom B and Chess, Benjamin and Child, Rewon and Gray, Scott and Radford, Alec and Wu, Jeffrey and Amodei, Dario},
  journal={arXiv preprint arXiv:2001.08361},
  year={2020}
}

@inproceedings{xie2023doremi,
  title={{DoReMi}: Optimizing Data Mixtures Speeds Up Language Model Pretraining},
  author={Xie, Sang Michael and Pham, Hieu and Dong, Xuanyi and Du, Nan and Liu, Hanxiao and Lu, Yifeng and Liang, Percy and Le, Quoc V. and Ma, Tengyu and Yu, Adams Wei},
  booktitle={Advances in Neural Information Processing Systems},
  volume={36},
  pages={69798--69818},
  year={2023}
}

@inproceedings{liu2025regmix,
title={RegMix: Data Mixture as Regression for Language Model Pre-training},
author={Qian Liu and Xiaosen Zheng and Niklas Muennighoff and Guangtao Zeng and Longxu Dou and Tianyu Pang and Jing Jiang and Min Lin},
booktitle={The Thirteenth International Conference on Learning Representations},
year={2025},
url={https://openreview.net/forum?id=5BjQOUXq7i}
}

@article{rubin2005causal,
  title={Causal inference using potential outcomes: Design, modeling, decisions},
  author={Rubin, Donald B},
  journal={Journal of the American statistical Association},
  volume={100},
  number={469},
  pages={322--331},
  year={2005},
  publisher={Taylor \& Francis}
}

@book{imbens2015causal,
  title={Causal Inference for Statistics, Social, and Biomedical Sciences: An Introduction},
  author={Imbens, Guido W and Rubin, Donald B},
  year={2015},
  publisher={Cambridge university press}
}

@inproceedings{fan2024doge,
  title={DOGE: Domain Reweighting with Generalization Estimation},
  author={Fan, Simin and Pagliardini, Matteo and Jaggi, Martin},
  booktitle={International Conference on Machine Learning},
  pages={12895--12915},
  year={2024},
  organization={PMLR}
}

@inproceedings{
ye2025data,
title={Data Mixing Laws: Optimizing Data Mixtures by Predicting Language Modeling Performance},
author={Jiasheng Ye and Peiju Liu and Tianxiang Sun and Jun Zhan and Yunhua Zhou and Xipeng Qiu},
booktitle={The Thirteenth International Conference on Learning Representations},
year={2025},
url={https://openreview.net/forum?id=jjCB27TMK3}
}

@inproceedings{
chen2025aioli,
title={Aioli: A Unified Optimization Framework for Language Model Data Mixing},
author={Mayee F Chen and Michael Y. Hu and Nicholas Lourie and Kyunghyun Cho and Christopher Re},
booktitle={The Thirteenth International Conference on Learning Representations},
year={2025},
url={https://openreview.net/forum?id=sZGZJhaNSe}
}

@inproceedings{renduchintala-etal-2024-smart,
    title = "{SMART}: Submodular Data Mixture Strategy for Instruction Tuning",
    author = "Renduchintala, H S V N S Kowndinya  and
      Bhatia, Sumit  and
      Ramakrishnan, Ganesh",
    editor = "Ku, Lun-Wei  and
      Martins, Andre  and
      Srikumar, Vivek",
    booktitle = "Findings of the Association for Computational Linguistics: ACL 2024",
    month = aug,
    year = "2024",
    address = "Bangkok, Thailand",
    publisher = "Association for Computational Linguistics",
    url = "https://aclanthology.org/2024.findings-acl.766/",
    doi = "10.18653/v1/2024.findings-acl.766",
    pages = "12916--12934"
}

@inproceedings{li2025data,
  title={Data Mixing Optimization for Supervised Fine-Tuning of Large Language Models},
  author={Li, Yuan and Liu, Zhengzhong and Xing, Eric},
  booktitle={Proceedings of the 42nd International Conference on Machine Learning},
  pages={35419--35437},
  volume={267},
  series={Proceedings of Machine Learning Research},
  publisher={PMLR},
  year={2025},
  url={https://proceedings.mlr.press/v267/li25bh.html}
}

@inproceedings{
ming2026ideal,
title={{IDEAL}: Data Equilibrium Adaptation for Multi-Capability Language Model Alignment},
author={Chenlin Ming and Chendi Qu and Qizhi Pei and Zhuoshi Pan and Yu Li and Xiaoming Duan and Lijun Wu and Conghui He},
booktitle={The Fourteenth International Conference on Learning Representations},
year={2026},
url={https://openreview.net/forum?id=n9wS0Hdvri}
}

@book{peters2017elements,
  title={Elements of causal inference: foundations and learning algorithms},
  author={Peters, Jonas and Janzing, Dominik and Scholkopf, Bernhard},
  year={2017},
  publisher={MIT press}
}

@article{scholkopf2021toward,
  title={Toward causal representation learning},
  author={Sch{\"o}lkopf, Bernhard and Locatello, Francesco and Bauer, Stefan and Ke, Nan Rosemary and Kalchbrenner, Nal and Goyal, Anirudh and Bengio, Yoshua},
  journal={Proceedings of the IEEE},
  volume={109},
  number={5},
  pages={612--634},
  year={2021},
  publisher={IEEE}
}

@book{pearl2009causality,
  title={Causality},
  author={Pearl, Judea},
  year={2009},
  publisher={Cambridge university press}
}

@book{spirtes2000causation,
  title={Causation, prediction, and search},
  author={Spirtes, Peter and Glymour, Clark N and Scheines, Richard},
  edition={2},
  year={2001},
  publisher={MIT Press}
}

@inproceedings{shalit2017estimating,
  title={Estimating individual treatment effect: generalization bounds and algorithms},
  author={Shalit, Uri and Johansson, Fredrik D and Sontag, David},
  booktitle={International conference on machine learning},
  pages={3076--3085},
  year={2017},
  organization={PMLR}
}

@article{louizos2017causal,
  title={Causal effect inference with deep latent-variable models},
  author={Louizos, Christos and Shalit, Uri and Mooij, Joris M and Sontag, David and Zemel, Richard and Welling, Max},
  journal={Advances in neural information processing systems},
  volume={30},
  year={2017}
}

@article{shi2019adapting,
  title={Adapting neural networks for the estimation of treatment effects},
  author={Shi, Claudia and Blei, David and Veitch, Victor},
  journal={Advances in neural information processing systems},
  volume={32},
  year={2019}
}

@article{chernozhukov2018double,
  title={Double/debiased machine learning for treatment and structural parameters},
  author={Chernozhukov, Victor and Chetverikov, Denis and Demirer, Mert and Duflo, Esther and Hansen, Christian and Newey, Whitney and Robins, James},
  journal={The Econometrics Journal},
  volume={21},
  number={1},
  pages={C1--C68},
  year={2018},
  doi={10.1111/ectj.12097}
}

@article{zheng2018dags,
  title={Dags with no tears: Continuous optimization for structure learning},
  author={Zheng, Xun and Aragam, Bryon and Ravikumar, Pradeep K and Xing, Eric P},
  journal={Advances in neural information processing systems},
  volume={31},
  year={2018}
}

@inproceedings{
bengiometa,
title={A Meta-Transfer Objective for Learning to Disentangle Causal Mechanisms},
author={Yoshua Bengio and Tristan Deleu and Nasim Rahaman and Nan Rosemary Ke and Sebastien Lachapelle and Olexa Bilaniuk and Anirudh Goyal and Christopher Pal},
booktitle={International Conference on Learning Representations},
year={2020},
url={https://openreview.net/forum?id=ryxWIgBFPS}
}

@article{rojas2018invariant,
  title={Invariant models for causal transfer learning},
  author={Rojas-Carulla, Mateo and Sch{\"o}lkopf, Bernhard and Turner, Richard and Peters, Jonas},
  journal={Journal of Machine Learning Research},
  volume={19},
  number={36},
  pages={1--34},
  year={2018}
}

@article{arjovsky2019invariant,
  title={Invariant risk minimization},
  author={Arjovsky, Martin and Bottou, L{\'e}on and Gulrajani, Ishaan and Lopez-Paz, David},
  journal={arXiv preprint arXiv:1907.02893},
  year={2019}
}

@article{liu2021towards,
  title={Towards out-of-distribution generalization: A survey},
  author={Liu, Jiashuo and Shen, Zheyan and He, Yue and Zhang, Xingxuan and Xu, Renzhe and Yu, Han and Cui, Peng},
  journal={arXiv preprint arXiv:2108.13624},
  year={2021}
}


\appendix
\section{Experimental details}
\label{app:exp_details}

\subsection{Datasets}

We evaluate \method on two SFT datasets.

\textbf{\texttt{tulu-3-sft-mixture}}~\citep{lambert2025tulu} is used to train the Tulu 3 series of models. It contains 939,344 samples spanning seven domains: General (\texttt{Tulu 3 Hardcoded}, \texttt{OpenAssistant Guanaco}~\citep{pf2023openassistant}, \texttt{No Robots}~\citep{no_robots}, and \texttt{WildChat GPT-4}~\citep{zhao2024wildchat,deng2024wildvisopensourcevisualizer}), Knowledge Recall (\texttt{FLAN v2}~\citep{weifinetuned}, \texttt{SciRIFF}~\citep{wadden-etal-2025-sciriff}, and \texttt{TableGPT}~\citep{10.1145/3654979}), Math Reasoning (\texttt{Tulu 3 Persona MATH}, \texttt{Tulu 3 Persona GSM}, \texttt{Tulu 3 Persona Algebra}, \texttt{OpenMathInstruct 2}~\citep{toshniwal2025openmathinstruct}, and \texttt{NuminaMath-TIR}~\citep{numina_math_7b}), Coding (\texttt{Tulu 3 Persona Python} and \texttt{Evol CodeAlpaca}~\citep{luo2024wizardcoder}), Safety \& Non-Compliance (\texttt{CoCoNot}~\citep{brahman-kumar2024}, \texttt{Tulu 3 WildJailbreak}~\citep{wildteaming2024}, and \texttt{Tulu 3 WildGuardMix}~\citep{han2024wildguard}), Multilingual (\texttt{Aya}~\citep{singh2024aya}), and Precise IF (\texttt{Tulu 3 Persona IF}). We exclude the multilingual subset in our experiments.

\textbf{\texttt{AM-Thinking-v1-Distilled}}~\citep{tian2025correctanswersequaldistillation} is a reasoning dataset distilled from \texttt{AM-Thinking-v1}~\citep{ji2025amthinkingv1advancingfrontierreasoning}. It contains high-quality, automatically verified responses generated from a shared set of 1.89 million queries spanning a wide range of reasoning domains. Its format and verification pipeline allow for direct comparison and seamless integration into downstream tasks. It is intended to support the development of open-source language models with strong reasoning abilities. In our experiments, we use the code and math subsets.

We obtain the data-state covariates from \texttt{OpenDataArena-scored-data-2603}~\citep{opendataarena_tool_2025,cai2025opendataarena,opendataarena_scored_data_2025}.

\textbf{\texttt{OpenDataArena-scored-data-2603}}~\citep{opendataarena_tool_2025,cai2025opendataarena,opendataarena_scored_data_2025} is a scored SFT dataset collection comprising 63 high-quality instruction-following datasets with nearly 25 million samples. Its core value lies in its 30-dimensional scoring scheme: each sample is evaluated on metrics such as \texttt{Normalized\_Loss}~\citep{shum2025predictive}, \texttt{Writing\_Style}~\citep{10.5555/3692070.3694241}, \texttt{HES}~\citep{li2026unified}, and 27 others, enabling fine-grained data selection for filtering, curriculum learning, and mixture optimization.

\subsection{Models}

We use \texttt{Qwen2.5-0.5B} as the proxy model, scale the learned mixture strategy up to \texttt{Qwen2.5-7B}, and further conduct an extension experiment on \texttt{Qwen3-4B-Base}.

\textbf{\texttt{Qwen2.5}}~\citep{qwen2.5} is a series of large language models developed by Qwen. It includes both base and instruction-tuned models ranging from 0.5B to 72B parameters. Compared with \texttt{Qwen2}~\citep{qwen2}, \texttt{Qwen2.5} offers substantially more knowledge and stronger coding and mathematical reasoning capabilities, partly due to specialized expert models in these domains. It also improves instruction following, long-form generation (over 8K tokens), structured-data understanding (e.g., tables), and structured output generation, especially in JSON format. In addition, it is more robust to diverse system prompts, which improves role-play and controllability in chatbot settings. \texttt{Qwen2.5} supports contexts of up to 128K tokens and can generate up to 8K tokens, and it supports more than 29 languages, including Chinese, English, French, Spanish, Portuguese, German, Italian, Russian, Japanese, Korean, Vietnamese, Thai, and Arabic. In our experiments, we use the smallest model, \texttt{Qwen2.5-0.5B}, as the proxy and scale to the widely used \texttt{Qwen2.5-7B}.

\textbf{\texttt{Qwen3}}~\citep{qwen3technicalreport} is a newer generation of large language models than \texttt{Qwen2.5} in the Qwen series, offering a comprehensive suite of dense and mixture-of-experts (MoE) models. Built on extensive pretraining, \texttt{Qwen3} provides substantial advances in reasoning, instruction following, agent capabilities, and multilingual support. Its key features include seamless switching between a thinking mode for complex reasoning, mathematics, and coding and a non-thinking mode for efficient general-purpose dialogue within a single model, enabling strong performance across a wide range of scenarios. It also substantially improves reasoning performance, surpassing previous QwQ~\citep{qwq32b} models in thinking mode and \texttt{Qwen2.5}-Instruct models in non-thinking mode on mathematics, code generation, and commonsense reasoning. In addition, \texttt{Qwen3} shows stronger human preference alignment, with better performance in creative writing, role-playing, multi-turn dialogue, and instruction following, resulting in a more natural and engaging conversational experience. It also offers strong agent capabilities, enabling effective integration with external tools in both thinking and non-thinking modes and achieving leading performance among open-source models on complex agentic tasks. Finally, it supports more than 100 languages and dialects and demonstrates strong multilingual instruction-following and translation capabilities. In our extension experiments, we use the 4B dense model.

\subsection{Benchmarks}

Following Tulu 3, we assess model performance on multiple tasks and corresponding benchmarks, including Knowledge (MMLU~\citep{hendrycks2021measuring}, MMLU-Pro~\citep{wang2024mmlupro}, GPQA~\citep{rein2024gpqa}), Reasoning (BBH~\citep{srivastava2023beyond}, DROP~\citep{dua-etal-2019-drop}, AGIEval~\citep{zhong-etal-2024-agieval}), Math (GSM8K~\citep{cobbe2021trainingverifierssolvemath}, MATH~\citep{math}, OlympiadBench~\citep{he-etal-2024-olympiadbench}), Code (HumanEval~\citep{chen2021evaluatinglargelanguagemodels}, HumanEval+~\citep{liu2023is}, MBPP~\citep{austin2021programsynthesislargelanguage}), Instruction Following (IFEval~\citep{zhou2023instruction}, IFBench~\citep{pyatkin2026generalizing}), and Safety (ToxiGen~\citep{hartvigsen-etal-2022-toxigen}, TruthfulQA~\citep{lin-etal-2022-truthfulqa}).

\textbf{MMLU}~\citep{hendrycks2021measuring} is heterogeneous with respect to the reasoning skills required to answer the questions, including instances that require basic factual recall as well as those that demand logical reasoning and problem-solving skills. Following Tulu 3, we use a zero-shot chain-of-thought (CoT) setting that asks the model to ``summarize'' its reasoning before answering the question. We compute the macro average over all subjects in MMLU as the final task metric.

\textbf{MMLU-Pro}~\citep{wang2024mmlupro} is a 10-way multiple-choice extension of the MMLU dataset. We use essentially the same prompt and answer extraction procedure as in our AGIEval setup, adjusting only the number of answer choices.

\textbf{GPQA}~\citep{rein2024gpqa} is a set of very challenging multiple-choice questions written by domain experts in biology, physics, and chemistry. We use the same zero-shot prompt and answer extraction procedure as for AGIEval.

\textbf{BBH (BigBench-Hard)}~\citep{srivastava2023beyond} contains challenging reasoning problems for which models benefit from step-by-step reasoning. We use the default setting of OpenCompass~\citep{2023opencompass}.

\textbf{DROP}~\citep{dua-etal-2019-drop} is a reading comprehension task that requires discrete reasoning. We use the default setting of OpenCompass.

\textbf{AGIEval (English subset)}~\citep{zhong-etal-2024-agieval} includes the English-language subset of the AGIEval benchmark, specifically the following multiple-choice tasks: \textit{aqua-rat}, \textit{logiqa-en}, \textit{lsat-ar}, \textit{lsat-lr}, \textit{lsat-rc}, \textit{sat-en}, \textit{sat-math}, and \textit{gaokao-english}. We formulate the task using a simple zero-shot CoT prompt that encourages concise reasoning ending with a clearly stated answer choice. The model's answer choice is extracted by first matching the requested format, with fallback patterns if the format is not followed precisely. Specifically, we first look for the exact phrase indicated in the prompt (``\texttt{Therefore, the answer is [ANSWER]}'') and take the last such match. If that fails, we look for a sequence of softer variants, such as ``\texttt{answer is [ANSWER]}'' or ``\texttt{answer: [ANSWER]}'', before falling back to the last parenthesized letter found; if that also fails, we use the last stand-alone capital letter.

\textbf{GSM8K}~\citep{cobbe2021trainingverifierssolvemath} contains grade-school math word problems. We use the default setting of OpenCompass.

\textbf{MATH}~\citep{math} contains problems from mathematics competitions spanning various categories, such as algebra and calculus. We use the default setting of OpenCompass. We compute the macro average across subsections to obtain the final task metric.

\textbf{OlympiadBench}~\citep{he-etal-2024-olympiadbench} is an Olympiad-level bilingual multimodal scientific benchmark featuring 8,476 problems from Olympiad-level mathematics and physics competitions, including the Chinese college entrance exam. We evaluate only the English math subset and use the same evaluation logic as for MATH.

\textbf{HumanEval}~\citep{chen2021evaluatinglargelanguagemodels} and \textbf{HumanEval+}~\citep{liu2023is} evaluate models' ability to complete Python code from docstrings. HumanEval+ uses a more rigorous evaluation procedure than the original HumanEval benchmark, with additional tests. We use the default setting of OpenCompass.

\textbf{MBPP}~\citep{austin2021programsynthesislargelanguage} contains 974 programming tasks designed to be solvable by entry-level programmers. We use the default setting of OpenCompass.

\textbf{IFEval}~\citep{zhou2023instruction} evaluates the instruction-following ability of models in a setting where each instruction corresponds to constraints such that it can be programmatically verified whether the outputs satisfy those constraints. We use the default setting of OpenCompass and measure prompt-level accuracy in the loose evaluation setting.

\textbf{IFBench}~\citep{pyatkin2026generalizing} is designed to evaluate generalization in precise instruction following on 58 new, diverse, and challenging verifiable out-of-domain constraints. We use the default setting of OpenCompass.

\textbf{ToxiGen}~\citep{hartvigsen-etal-2022-toxigen} is a large-scale machine-generated dataset of 274k toxic and benign statements about 13 minority groups. We use a zero-shot setting with unnormalized accuracy.

\textbf{TruthfulQA}~\citep{lin-etal-2022-truthfulqa} is a benchmark for measuring whether a language model generates truthful answers to questions. The benchmark comprises 817 questions spanning 38 categories, including health, law, finance, and politics. We use the test split of \texttt{mc1} in a zero-shot setting.

We further partition these benchmarks into the development set $\mathcal{S}_{\mathrm{Dev}}$ and the unseen set $\mathcal{S}_{\mathrm{Uns}}$. $\mathcal{S}_{\mathrm{Dev}}$ comprises MMLU, MMLU-Pro, BBH, DROP, GSM8K, MATH, HumanEval, MBPP, IFEval, and TruthfulQA, while $\mathcal{S}_{\mathrm{Uns}}$ consists of GPQA, AGIEval, OlympiadBench, HumanEval+, IFBench, and ToxiGen.

\subsection{Baselines}

We compare \method with recent methods for offline data mixture optimization.

\textbf{RegMix}~\citep{liu2025regmix} is designed to automatically identify a high-performing data mixture by formulating the problem as a regression task. It trains many small models on diverse data mixtures, uses regression to predict the performance of unseen mixtures, and applies the best predicted mixture to train a large-scale model with orders of magnitude more compute.

\textbf{DoReMi}~\citep{xie2023doremi} first trains a small proxy model using group distributionally robust optimization (Group DRO) over domains to obtain domain weights (mixture proportions) without access to downstream tasks. It then resamples the dataset according to these domain weights and trains a larger full-sized model.

\textbf{ODM}~\citep{albalak2023efficientonlinedatamixing} combines elements of both data selection and data mixing. Based on multi-armed bandit algorithms, ODM optimizes the data mixing proportions during training.

\textbf{DMO}~\citep{li2025data} frames data mixing as an optimization problem and introduces a method designed to minimize validation loss. DMO parameterizes the loss by modeling effective data transfer and leveraging scaling laws for fine-tuning.

\subsection{Computing costs}

We train 512 proxy models with 0.5B parameters, each on 100K SFT examples. The average sequence length is approximately 4096 tokens, and the total estimated FLOPs are $5.53 \times 10^{20}$. Because our method is state-aware, it maintains strong generalization when transferred to out-of-distribution (OOD) data, as demonstrated by the extension experiments in Section~\ref{sec:extens_exp}, without requiring the proxy models to be retrained.

For fair comparison, we use the same proxy-model configuration for baseline methods that require proxy training. Specifically, for RegMix~\citep{liu2025regmix}, we also use 512 proxy models. For DoReMi~\citep{xie2023doremi}, we use a single proxy model. For ODM~\citep{albalak2023efficientonlinedatamixing}, we use a single model to determine the data mixing proportions and training order. For DMO~\citep{li2025data}, since it is trained on the same data, we directly use the mixture proportions reported in the original paper.

\subsection{Hyperparameters}

All random seeds in our experiments are set to 42, and all experiments are conducted on NVIDIA H800 GPUs. 

\paragraph{Training.} For \texttt{Qwen2.5-0.5B} and \texttt{Qwen2.5-7B}, we follow DMO~\citep{li2025data}; for \texttt{Qwen3-4B-Base}, we follow OpenDataArena~\citep{cai2025opendataarena}. All training hyperparameters are listed in Table~\ref{tab:hyper_train}.

\paragraph{Evaluation.} All evaluation hyperparameters are listed in Table~\ref{tab:hyper_eval}. For \texttt{Qwen2.5-0.5B} and \texttt{Qwen2.5-7B}, we set \texttt{max\_tokens} to 4096, whereas for \texttt{Qwen3-4B-Base}, we set it to 32,768. This difference is determined by whether the training data includes LongCoT-style reasoning.

\begin{table}[!htb]
    \centering
    \caption{Training hyperparameters for \texttt{Qwen2.5-0.5B}, \texttt{Qwen2.5-7B} and \texttt{Qwen3-4B}.}
    \label{tab:hyper_train}
    \begin{minipage}{0.32\textwidth}
        \centering
        \resizebox{\textwidth}{!}{\begin{tabular}{lc}
            \toprule
            \textbf{Hyperparameter} & \textbf{Value} \\
            \midrule
            \rowcolor{blue!15}\multicolumn{2}{c}{\texttt{Qwen2.5-0.5B}} \\
            \texttt{learning\_rate} & \texttt{2.0e-5} \\
            \texttt{num\_train\_epochs} & \texttt{3} \\
            \texttt{num\_gpus} & $8$ \\
            \texttt{per\_device\_train\_batch\_size} & \texttt{32} \\
            \texttt{gradient\_accumulation\_steps} & \texttt{1} \\
            \texttt{lr\_scheduler\_type} & \texttt{cosine} \\
            \texttt{warmup\_ratio} & \texttt{0.1} \\
            \texttt{cutoff\_len} & \texttt{4096} \\
            \texttt{deepspeed} & \texttt{z0} \\
            \texttt{flash\_attn} & \texttt{fa2} \\
            \texttt{use\_liger\_kernel} & \texttt{true} \\
            \texttt{bf16} & \texttt{true} \\
            \bottomrule
        \end{tabular}}
    \end{minipage}
    \begin{minipage}{0.32\textwidth}
        \centering
        \resizebox{\textwidth}{!}{\begin{tabular}{lc}
            \toprule
            \textbf{Hyperparameter} & \textbf{Value} \\
            \midrule
            \rowcolor{blue!15}\multicolumn{2}{c}{\texttt{Qwen2.5-7B}} \\
            \texttt{learning\_rate} & \texttt{5.0e-6} \\
            \texttt{num\_train\_epochs} & \texttt{3} \\
            \texttt{num\_gpus} & $8$ \\
            \texttt{per\_device\_train\_batch\_size} & \texttt{16} \\
            \texttt{gradient\_accumulation\_steps} & \texttt{2} \\
            \texttt{lr\_scheduler\_type} & \texttt{cosine} \\
            \texttt{warmup\_ratio} & \texttt{0.1} \\
            \texttt{cutoff\_len} & \texttt{4096} \\
            \texttt{deepspeed} & \texttt{z2} \\
            \texttt{flash\_attn} & \texttt{fa2} \\
            \texttt{use\_liger\_kernel} & \texttt{true} \\
            \texttt{bf16} & \texttt{true} \\
            \bottomrule
        \end{tabular}}
    \end{minipage}
    \begin{minipage}{0.32\textwidth}
        \centering
        \resizebox{\textwidth}{!}{\begin{tabular}{lc}
            \toprule
            \textbf{Hyperparameter} & \textbf{Value} \\
            \midrule
            \rowcolor{blue!15}\multicolumn{2}{c}{\texttt{Qwen3-4B}} \\
            \texttt{learning\_rate} & \texttt{5.0e-5} \\
            \texttt{num\_train\_epochs} & \texttt{3} \\
            \texttt{num\_gpus} & $8$ \\
            \texttt{per\_device\_train\_batch\_size} & \texttt{2} \\
            \texttt{gradient\_accumulation\_steps} & \texttt{2} \\
            \texttt{lr\_scheduler\_type} & \texttt{cosine} \\
            \texttt{warmup\_ratio} & \texttt{0.1} \\
            \texttt{cutoff\_len} & \texttt{32768} \\
            \texttt{deepspeed} & \texttt{z2} \\
            \texttt{flash\_attn} & \texttt{fa2} \\
            \texttt{use\_liger\_kernel} & \texttt{true} \\
            \texttt{bf16} & \texttt{true} \\
            \bottomrule
        \end{tabular}}
    \end{minipage}
\end{table}

\begin{table}[!htb]
    \centering
    \caption{Evaluation hyperparameters.}
    \label{tab:hyper_eval}
    \begin{tabular}{lc}
    \toprule
    \textbf{Hyperparameter} & \textbf{Value} \\
    \midrule
    \texttt{pass@n} & \texttt{n=1} \\
    \texttt{presence\_penalty} & \texttt{0.0} \\
    \texttt{frequency\_penalty} & \texttt{0.0} \\
    \texttt{repetition\_penalty} & \texttt{1.0} \\
    \texttt{temperature} & \texttt{0.0} \\
    \texttt{top\_p} & \texttt{1.0} \\
    \texttt{top\_k} & \texttt{-1} \\
    \texttt{min\_p} & \texttt{0.0} \\
    \texttt{max\_tokens} & \texttt{4096} / \texttt{32768} \\
    \texttt{min\_tokens} & \texttt{0} \\
    \bottomrule
    \end{tabular}
\end{table}

\section{Proof of the analytical mixture policy}
\label{app:analytical_proof}

In this section, we provide a rigorous mathematical derivation for the analytical policy extraction \method-A described in Section~\ref{sec:policy_extraction}. 

Given the estimated state-conditioned marginal return $\hat{\theta}(X_{\mathrm{tar}}) \in \mathbb{R}^K$, our objective is to find the optimal raw mixture strategy $T^*$ that maximizes the expected causal performance gain under the Level-Log formulation. This yields a constrained optimization problem over the probability simplex:
\[
\max_{T} \quad \mathcal{J}(T) = \sum_{k=1}^{K} \hat{\theta}_k \log(T_k)
\]
\[
\text{subject to} \quad \sum_{k=1}^{K} T_k = 1, \quad T_k \geq 0 \quad \forall k \in \{1, \dots, K\}.
\]

To explicitly handle the non-negativity constraints, we reformulate the problem as a minimization problem and apply the Karush--Kuhn--Tucker (KKT) conditions. We minimize $-\mathcal{J}(T)$ and define the Lagrangian $\mathcal{L}(T, \lambda, \mu)$:
\[
\mathcal{L}(T, \lambda, \mu) = - \sum_{k=1}^{K} \hat{\theta}_k \log(T_k) + \lambda \left( \sum_{k=1}^{K} T_k - 1 \right) - \sum_{k=1}^{K} \mu_k T_k,
\]
where $\lambda \in \mathbb{R}$ is the Lagrange multiplier for the equality constraint, and $\mu_k \geq 0$ are the KKT multipliers for the inequality constraints $T_k \geq 0$.

The KKT optimality conditions require that the optimal solution $(T^*, \lambda^*, \mu^*)$ satisfy: (1) Stationarity: $\frac{\partial \mathcal{L}}{\partial T_k} = -\frac{\hat{\theta}_k}{T_k^*} + \lambda^* - \mu_k^* = 0$, which implies $\lambda^* - \mu_k^* = \frac{\hat{\theta}_k}{T_k^*}$. (2) Primal feasibility: $\sum_{k=1}^{K} T_k^* = 1$ and $T_k^* \geq 0$. (3) Dual feasibility: $\mu_k^* \geq 0$. (4) Complementary slackness: $\mu_k^* T_k^* = 0$.

We analyze the optimal solution by partitioning the domains based on the sign of their estimated marginal returns $\hat{\theta}_k$:

\paragraph{Domains with negative or zero marginal returns ($\hat{\theta}_k \leq 0$).} 
Suppose, for the sake of contradiction, that $T_k^* > 0$. By the complementary slackness condition, $T_k^* > 0 \implies \mu_k^* = 0$. Substituting this into the stationarity condition yields $\lambda^* = \frac{\hat{\theta}_k}{T_k^*}$. Since $\hat{\theta}_k \leq 0$ and $T_k^* > 0$, this implies $\lambda^* \leq 0$. However, there must exist at least one domain $j$ with $\hat{\theta}_j > 0$ and $T_j^* > 0$ (otherwise the objective is unbounded negatively, and empirical mixtures always contain positive-return domains). For that domain $j$, $\mu_j^* = 0$ implies $\lambda^* = \frac{\hat{\theta}_j}{T_j^*} > 0$, leading to a contradiction. Furthermore, since $\log(T_k) \to -\infty$ as $T_k \to 0$, a negative $\hat{\theta}_k$ pushes the objective value to $+\infty$ as $T_k \to 0^+$. Therefore, the optimal allocation strictly binds at the boundary:
\[
T_k^* = 0 \quad \text{for all} \quad \hat{\theta}_k \leq 0.
\]

\paragraph{Domains with positive marginal returns ($\hat{\theta}_k > 0$).}
Let $\mathcal{P} = \{k \mid \hat{\theta}_k > 0\}$ denote the active set. For $k \in \mathcal{P}$, since $T_k^* > 0$ (otherwise the objective drops to $-\infty$), the complementary slackness condition dictates $\mu_k^* = 0$. The stationarity condition simplifies to:
\[
\lambda^* = \frac{\hat{\theta}_k}{T_k^*} \implies T_k^* = \frac{\hat{\theta}_k}{\lambda^*}.
\]
To determine $\lambda^*$, we invoke the primal feasibility condition over the active set $\mathcal{P}$:
\[
\sum_{k \in \mathcal{P}} T_k^* = \sum_{k \in \mathcal{P}} \frac{\hat{\theta}_k}{\lambda^*} = 1 \implies \lambda^* = \sum_{k \in \mathcal{P}} \hat{\theta}_k.
\]
Substituting $\lambda^*$ back, we obtain the exact proportional assignment:
\[
T_k^* = \frac{\hat{\theta}_k}{\sum_{j \in \mathcal{P}} \hat{\theta}_j} \quad \text{for} \quad k \in \mathcal{P}.
\]

By unifying both cases, the global optimal solution maps strictly to zero for non-positive causal effects, and scales proportionally for positive effects. This is analytically identical to applying a Rectified Linear Unit (ReLU) activation to the causal marginal returns followed by $L_1$ normalization:
\[
T_k^{\mathrm{A}} = \frac{[\hat{\theta}_k(X_{\mathrm{tar}})]_+}{\sum_{j=1}^{K} [\hat{\theta}_j(X_{\mathrm{tar}})]_+}, \qquad \text{where} \quad [a]_+ = \max(a, 0).
\]
This completes the proof, confirming that our analytical extraction is the mathematically exact closed-form policy under the simplex constraint.


\end{document}